\documentclass{article}

\PassOptionsToPackage{numbers,compress}{natbib}
\usepackage[final]{Styles/neurips_2025}

\usepackage[utf8]{inputenc}      
\usepackage[T1]{fontenc}         
\usepackage{microtype}           
\usepackage{xcolor}              
\usepackage{url}                 

\usepackage{amsmath,amssymb,amsthm,mathtools,bm}  

\usepackage{booktabs}            
\usepackage{graphicx}            
\usepackage{adjustbox}
\usepackage{multirow}
\usepackage{subcaption}          

\usepackage{algorithm}           
\usepackage{algpseudocode}       
\usepackage{setspace}

\usepackage{hyperref}            
\usepackage[capitalize,noabbrev]{cleveref}  
\crefname{equation}{}{}          

\usepackage{abbreviations}       
\usepackage{enumitem}            

\usepackage[toc,page,header]{appendix}
\usepackage{minitoc}
\doparttoc                      
\faketableofcontents            

\title{Exploration via Feature Perturbation \\ in Contextual Bandits}

\author{
  Seouh-won Yi\\
  Seoul National University\\
  \texttt{uniqueseouh@snu.ac.kr}
  \And
  Min-hwan Oh\\
  Seoul National University\\
  \texttt{minoh@snu.ac.kr}
}

\begin{document}

\maketitle

\begin{abstract}
We propose \textit{feature perturbation}, a simple yet effective exploration strategy for contextual bandits that injects randomness directly into feature inputs, instead of randomizing unknown parameters or adding noise to rewards.
Remarkably, this algorithm achieves $\otil(d\sqrt{T})$ worst-case regret bound for generalized linear contextual bandits, while avoiding the $\otil(d^{3/2}\sqrt{T})$ regret typical of existing randomized bandit algorithms.
Because our algorithm eschews parameter sampling, it is both computationally efficient and naturally extends to non-parametric or neural network models. 
We verify these advantages through empirical evaluations, demonstrating that feature perturbation not only  surpasses existing methods but also unifies strong practical performance with the near-optimal regret guarantees.
\end{abstract}

\section{Introduction} \label{sec:introduction}

Multi-armed bandits (MABs) provide the canonical model for sequential decision-making under uncertainty: at each round a decision-making agent selects one of several arms to maximize cumulative reward while balancing exploration and exploitation. However, classical MABs ignore side information that often accompanies decisions in practice. Contextual bandits address this limitation by allowing the agent to first observe contextual information and then choose an action tailored to that context---e.g., features of users and/or items inform which arm to pull. This contextualized formulation has become a pivotal framework in online learning and sequential decision-making, with a rich literature on algorithms and guarantees~\cite{abe1999associative, auer2002finite, li2010contextual, MAL-024, lattimore2020bandit}.

A widely studied formulation of contextual bandits is the \emph{(generalized) linear contextual bandit}, where the expected reward is modeled by a linear function~\cite{abe1999associative, auer2002finite, li2010contextual, chu2011contextual, abbasi2011improved} or, more generally, by a generalized linear model (GLM)~\cite{filippi2010parametric, li2017provably, jun2017scalable, lee2025unified}.
In both linear and GLM settings, deterministic methods based on \emph{optimism in the face of uncertainty} (OFU)~\cite{auer2002finite, li2010contextual, abbasi2011improved, li2017provably} and randomized approaches such as Thompson Sampling (TS)~\cite{chapelle2011empirical, agrawal2013thompson, abeille2017Linear} or Perturbed History Exploration (PHE)~\cite{kveton2020perturbed, kveton2020randomized, lee2024improved} have been extensively studied.
Notably, OFU-type algorithms achieve near-optimal regret of $\otil(d\sqrt{T})$ in linear contextual bandits (and likewise in GLM bandits~\citep{li2017provably}), yet often underperform compared to TS and PHE in practice. 
In contrast, randomized exploration methods typically exhibit superior empirical performance but suffer from sub-optimal theoretical guarantees: standard analyses confirm a regret bound of $\otil(d^{3/2}\sqrt{T})$~\cite{abeille2017Linear, agrawal2013thompson} in the frequentist (worst-case) setting.%
\footnote{LinTS~\cite{abeille2017Linear, agrawal2013thompson} achieves a regret of $\otil(\min(d^{3/2}\sqrt{T},d\sqrt{T\log K}))$. 
While \citet{kveton2020perturbed} originally showed that LinPHE has a regret of $\otil(d\sqrt{T\log K})$, where $K$ is the number of arms, a recent work \cite{lee2024improved} proves that LinPHE also satisfies $\otil(d^{3/2}\sqrt{T})$ regret.
For further discussion on trading off factors of $\Ocal(\sqrt{d})$ vs.$\Ocal(\sqrt{\log K})$, see \citet{agrawal2013thompson}.
In this work, we consider even a large action space with $K > e^d$, so the $\otil(d^{3/2}\sqrt{T})$ regret of randomized (generalized) linear bandit algorithms is the main focus of discussion.
}
Crucially, this gap is not merely an artifact of analysis: \citet{hamidi2020frequentist} show it reflects an inherent cost of randomization in (generalized) linear Thompson sampling. 
This result highlights a fundamental mismatch between the existing randomized exploration and the tighter optimism mechanism in OFU-based approaches.
This dichotomy prompts a natural question: \emph{is it possible to close the gap between randomized exploration and $\otil(d\sqrt{T})$ worst-case regret?} 
If one adheres strictly to randomly sampling the parameters (as in TS) or perturbing the observed rewards (as in PHE), there appears a fundamental barrier~\citep{hamidi2020frequentist} preventing regret from achieving $\otil(d\sqrt{T})$. 

In this work, we propose a simple yet effective alternative: instead of sampling parameters or perturbing rewards, we randomly perturb the observed \emph{features} (or contexts). 
By shifting the focus of exploration from parameter space to feature space, we circumvent the limitations that impose higher regret on existing randomized algorithms. 
Remarkably, our analysis shows that this new approach not only retains the empirical advantages of randomization but also achieves $\otil(d\sqrt{T})$ worst-case regret in (generalized) linear bandit settings, with no additional dependence on the number of arms. 
Furthermore, our method avoids the computational overhead of sampling parameters, making it attractive for a wide range of real-world applications.

Beyond theoretical efficiency, feature perturbation can seamlessly extend to more flexible or non-parametric reward models, including neural networks. 
We demonstrate this empirically, showing that feature-based randomization can drive effective exploration even when specific parametric model assumptions may not hold.
Hence, our proposed approach unifies strong theoretical guarantees with practical efficacy in (generalized) linear contextual bandits, and extends practically to more complex models.
Our main contributions are summarized as follows:
\begin{itemize}
    \item \textbf{Feature perturbation for contextual bandits.} We introduce a \emph{new class of algorithms} for randomized exploration, termed \emph{feature perturbation}, which focuses on perturbing feature inputs rather than parameters or rewards. This approach is straightforward to implement and conceptually distinct from existing randomized exploration strategies.

    \item \textbf{Tight regret bounds.} To the best of our knowledge, our work is the first \emph{randomized algorithm} for generalized linear contextual bandits that achieves: (i) a regret bound of $\otil(d\sqrt{T})$, matching the best-known guarantees of deterministic (OFU-based) methods; and simultaneously (ii) benefiting from an instance-dependent constant $\kappa$. Notably, our algorithm’s regret does not increase (even logarithmically) with the number of arms.

    \item \textbf{Empirical validation.} Through extensive experiments on both synthetic and real-world data, we show that feature perturbation not only performs competitively against existing randomized methods but also generalizes beyond parametric models (e.g., deep neural networks), demonstrating robustness even when linear assumptions do not hold.
\end{itemize}
\section{Related works}
Contextual bandits have been extensively investigated under various modeling assumptions. 
In the \emph{linear} bandit setting, deterministic methods based on OFU~\cite{abbasi2011improved, auer2002finite} achieve near-optimal $\widetilde{O}(d\sqrt{T})$ regret, but often exhibit conservative exploration in practice.
By contrast, \emph{randomized} algorithms such as TS~\cite{abeille2017Linear, agrawal2013thompson, chapelle2011empirical} and PHE~\cite{kveton2020perturbed, kveton2020randomized} typically show better empirical performance yet suffer from a higher $\widetilde{O}(d^{3/2}\sqrt{T})$ regret bound. 
Notably, \citet{hamidi2020frequentist} demonstrated that the extra $\sqrt{d}$ inflation in TS-type algorithms is unavoidable in worst-case scenarios: eliminating this factor would lead to a linear dependence on $T$.
Consequently, parameter-based randomization cannot, in general, achieve $\widetilde{O}(d\sqrt{T})$ regret without further modifications.

Generalized linear bandits (GLB; \citep{filippi2010parametric, li2017provably}) extend linear bandits to settings where rewards follow a nonlinear link function. 
UCB- and TS-based approaches~\citep{abeille2017Linear, filippi2010parametric, kveton2020randomized, vaswani20old} have also been applied here, displaying the same contrast between deterministic and randomized exploration.
While UCB-type methods reach $\widetilde{O}(d\sqrt{T})$ regret, they tend to over-explore in practice; randomized strategies mitigate this over-exploration but retain an additional $\sqrt{d}$ penalty in the worst case. 
Like their linear counterparts, these methods rely on sampling the unknown parameter or perturbing rewards rather than altering the feature representation.

By contrast, our work introduces a new class of \emph{feature-perturbation} (FP) algorithms designed to circumvent the dimensional penalty inherent in standard randomized approaches. Instead of randomizing parameters or rewards, we propose to perturb the features directly.
This perspective not only preserves the empirical robustness associated with randomized strategies but also achieves a tight regret bound in both linear and generalized linear settings---thereby reconciling the theoretical and practical advantages of contextual bandit exploration.
\section{Preliminaries} \label{sec:preliminaries}
\paragraph{Notations.} 
For vectors $x,y \in \RR^d$, let $\|x\|$ denote the $2$-norm and $\|x\|_A = \sqrt{x^\top A x}$ the weighted norm for a positive definite matrix $A \in \RR^{d \times d}$. 
The inner product is $x^\top y=\dotp{x}{y}$, and the weighted version is $x^\top Ay=\dotp{x}{y}_A$.  
The notation $\otil$ hides logarithmic factors in big-O notation, retaining instance-dependent constants.
For a real-valued function $f$, we write $\dot{f}$ and $\ddot{f}$ to denote its first and second derivatives.
The set $\cbr{1, \ldots, K}$ is abbreviated as $\sbr{K}$.

\paragraph{Generalized linear contextual bandit.}
A \emph{generalized linear model} (GLM; \cite{GLM}) describes a response $r\in\RR$ drawn from an exponential-family distribution with mean $\mu(x^\top\theta^*)$, where $x\in\RR^d$ is a feature vector and $\theta^*\in\RR^d$ is an unknown parameter.
Given differentiable functions $g$ and $h$, and a base measure $\nu$, the conditional density of $r$ given $x$ takes the form: 
\begin{equation}\label{eqn:density}
    dp(r\mid x;\theta^*) = \exp\rbr{rx^\top\theta^*-g(x^\top\theta^*)+h(r)} d\nu,
\end{equation}
where the derivative of $g$ defines the \textit{link function} $\mu$.\footnote{We normalize the reward model so that $\Var[r_t\mid x_t]=\dot{\mu}(x_t^\top\theta^*)$. Scaling the variance by $\sigma^2$ (by inserting $\sigma^2$ in the denominator of the exponential term of Eq.~\eqref{eqn:density}) accordingly yields a $\sigma$-inflated regret bound.} 
Let $\Hcal_{t-1} := \sigma(\cbr{(x_\tau, r_\tau)}_{\tau=1}^{t-1})$ denote the filtration up to round $t-1$.
We define $\PP_t(\cdot):=\PP(\cdot\mid\Hcal_{t-1})$ and $\EE_t\sbr{\cdot}:=\EE\sbr{\cdot\mid\Hcal_{t-1}}$. 
The negative log-likelihood and the \textit{maximum likelihood estimate} (MLE) at round $t$ are then given by:
\begin{equation*}
L_t(\theta) = \sum_{\tau=1}^{t-1} \big(g(x_\tau^\top\theta) - r_\tau x_\tau^\top\theta\big), \quad \htt := \argmin_{\theta \in \Theta} L_t(\theta).
\end{equation*}
In the generalized linear contextual bandit (GLB) setting, the agent observes a context $c_t \in \Ccal$ and a corresponding set of feature vectors $\Xcal_t\subset\RR^d$ representing each allowable arm $a\in\Acal(c_t)$ at each round. 
Upon selecting $x_t \in \Xcal_t$, the agent receives a stochastic reward $r_t\sim p(\cdot\mid x_t;\theta^*)$.
The learner aims to minimize the regret: $R(T) = \sum_{t=1}^T \rbr{\mu(x_{t*}^\top\theta^*) - \mu(x_t^\top\theta^*)}$, where $x_{t*} := \argmax_{x \in \Xcal_t} \mu(x^\top\theta^*)$ is the optimal arm at round $t$, which depends on the context $c_t$.
\section{Algorithm: \texorpdfstring{$\glmfp$}{GLM-FP}} \label{sec:algorithm}
\begin{algorithm}[htb]
\setstretch{1.2}
    \caption{$\glmfp$: Feature Perturbation in Generalized Linear Bandits}
\begin{algorithmic}[1]
    \State {\bfseries Input:} Regularization parameter $\lambda>0$, tuning parameter $\{c_t\}$
    \For{$t=1,2,\ldots,T$}
        \State Compute $\htt=\argmin_{\theta\in\RR^d}L_t(\theta;\{x_\tau,r_\tau\}_{\tau=1}^{t-1})$
        \State Sample $\zeta_t \sim \Ncal(\zero,\Ib)$
        \State Compute $\tx_{ti}=x_{ti}+c_t \cdot\frac{\|x_{ti}\|_{\hat{H}_t^{-1}}}{\|\htt\|}\cdot \zeta_t$ for all $i$
        \State Choose $i_t=\argmax_{i\in[|\Xcal_t|]}~\mu(\tx_{ti}^\top\htt)$ and observe reward $r_t$\Comment{Let $x_t:=x_{t,i_t}$}
    \EndFor
\end{algorithmic}
\label{Alg:GLM-FP}
\end{algorithm}
At each step $t$, given the filtration $\Hcal_{t-1}$, the algorithm computes the MLE $\htt$ (line 3). 
Since MLE lacks a closed-form solution, we employ \emph{Sequential Quadratic Programming} (SQP; \cite{SQP}) or \emph{Iteratively Reweighted Least Squares} (IRLS; \cite{IRLS}).
Instead of perturbing rewards or parameters, the algorithm injects controlled randomness into feature vectors using a perturbing distribution $\Dcal$. 
By default, we use a multivariate normal distribution (line 4), but any distribution satisfying concentration and anti-concentration properties can be employed, as analyzed in \cref{sec:conc anti}.

A fundamental property of the GLB is that the strictly increasing link function simplifies the problem structure, making it resemble linear bandits. 
While prior works~\citep{abeille2017Linear, li2017provably, vaswani20old} adopt near-identical methods to linear bandits---typically using the \emph{vanilla Gram matrix} $V_t=\lambda\Ib+\sum_{\tau=1}^{t-1}x_\tau x_\tau^\top$---our approach takes a more refined direction. 
To precisely control the perturbation magnitude for each feature representation, our approach utilizes a weighted Gram matrix, $\hat{H}_t:=\lambda\Ib+\nabla^2 L_t(\htt)$. 

This adaptive structure, modulated by a tunable parameter $c_t$, enables a more targeted perturbation strategy.
Using the resulting scaling factors, the algorithm perturbs each feature vector to construct a set of perturbed vectors, $\cbr{\tilde{x}_{ti}}$ for reachable arms at round $t$ (line 5).
Importantly, the perturbation noise $\zeta_t$ is shared across all arms, coupling each feature vector with the same random variable.
This design eliminates the explicit dependence on $K$ in the regret bound, thereby yielding stronger theoretical guarantees and improved empirical performance. 
Finally, the algorithm selects the arm that maximizes $\mu(\tilde{x}_{ti}^\top\hat{\theta}_t)$ and updates its history upon observing the reward $r_t$ (lines 6).

\subsection{Extension to general function class}
The algorithm extends naturally to more flexible function classes, as described in \cref{alg:FP}. 
Under the realizability assumption (i.e., $f^*\!\in\!\Fcal$), the estimate $\hat{f}_t\!\in\!\Fcal$ is obtained via a least squares oracle on the history $\Hcal_{t-1}$~\cite{beyonducb,FALCON, uccb}. 
Given the structural of the bandit reward model, a sampling distribution is then defined for each arm and used to construct a set of perturbed contexts $\cbr{\tilde{x}_{ti}}$.
The algorithm selects the arm that maximizes $\hat{f}_t(\tilde{x}_{ti})$, thereby balancing exploration and exploitation.

The proposed $\glmfp$ algorithm is an instance of this framework, characterized by two specific design choices: (i) a Gaussian sampling distribution $\Dcal(x_{ti}, \Sigma_{ti}) \triangleq \Ncal(x_{ti}, \Sigma_{ti})$ centered at $x_{ti}$ with elliptical covariance scaling with the uncertainty in the direction of $x_{ti}$ and normalized by the estimated parameter norm; and (ii) a \emph{coupled} perturbation scheme, where a single shared random vector $\zeta_t$ perturbs all arms simultaneously, in contrast to perturbing each arm independently.

\subsection{Intuition behind the algorithm}\label{sec:intuition}
In contextual bandit problems, randomized algorithms can be broadly categorized into two types: those that introduce randomness into the underlying model and those that inject randomness directly into the estimated expected rewards. 
The former category includes methods such as PHE~\cite{kveton2020perturbed} and TS~\cite{abeille2017Linear, agrawal2013thompson}, which compute perturbed model parameters to induce exploration. However, as noted by \citet{hamidi2020frequentist}, this approach can be suboptimal even in linear settings. The latter category includes algorithms such as RandUCB~\cite{vaswani20old}, where the model is trained deterministically and randomness is introduced by adding stochastic bonuses to the estimated rewards of each arm. While effective in inducing exploration, such reward perturbation can violate the inductive bias of the function class, since the resulting scores for arms may not be realized simultaneously by any single model $f \in \Fcal$—even in simple linear cases.

To address these limitations, we propose an alternative strategy that retains the estimated model $\hat{f}$ from past data and introduces randomness through input perturbations at decision time. 
Exploring directly in the feature space preserves the structural assumptions of the function class and remains effective even in overparameterized settings where $p \gg d$, such as neural bandits. 
It also aligns naturally with real-world scenarios in which contextual features contain inherent noise~\cite{Bastani2020online, Kannan2018greedy, Kim2023Noisy}.
\begin{figure*}[ht]
\begin{center}
\centerline{\includegraphics[width=\textwidth]{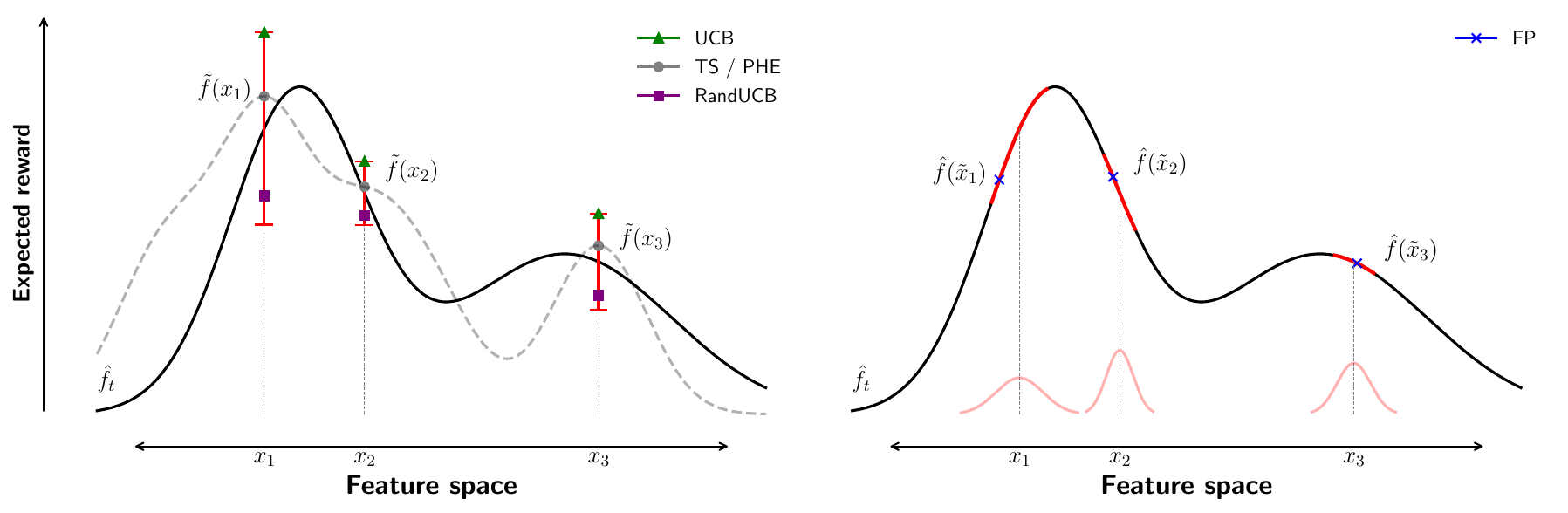}}
\caption[Comparison of exploration strategies.]{
(Left) Model perturbation methods randomize rewards via off-estimated models $\tilde{f}_t$.
(Right) Feature Perturbation (FP) perturbs inputs and evaluates them with a fixed model $\hat{f}_t$.}
\label{fig:explanation}
\end{center}
\vskip -0.3in
\end{figure*}

As illustrated in \cref{fig:explanation}, whereas UCB, TS, PHE, and RandUCB randomize parameters or reward estimates via modified models $\tilde{f}_t$, $\glmfp$ explores through feature-space perturbations evaluated under a fixed $\hat{f}$. 
This structural shift preserves the inductive bias of the model class and decouples exploration from parameter uncertainty, a property that becomes crucial in high-dimensional regimes.

\section{Regret analysis of \texorpdfstring{$\glmfp$}{GLM-FP}} \label{sec:analysis}
In this section, we establish the regret guarantee for the proposed algorithm, $\glmfp$, and outline the key steps in its proof. 
We introduce fundamental concepts and lemmas that form the backbone of our analysis. 
We begin by presenting the standard assumptions commonly adopted in the analysis of generalized linear bandit algorithms~\cite{abbasi2011improved, abeille2017Linear, Abeille2020InstanceWiseMA, agrawal2013thompson, Faury2020Improveds,filippi2010parametric, kveton2020randomized, lee2025unified, li2017provably, neuralucb}.
\begin{assumption}[Boundedness]\label{asm:Boundedness}
    The \emph{feature space} $\Xcal$ and the parameter space $\Theta$ are compact subsets of $\RR^d$. 
    For any $x\in\Xcal_{[T]}$ and $\theta^*\in\Theta$, we have $\|x\|\leq 1$ and $\|\theta^*\|\leq 1$.
\end{assumption}
\begin{assumption}[Self-concordance]\label{asm:Linkfunction}
    The function $g$ is three times differentiable, and its derivative $\dot{g}=\mu$ is strictly increasing and $\Lcal_\mu$-Lipschitz continuous. 
    Furthermore, $g$ is \emph{self-concordant}, characterized by the constant $M_\mu:=\sup_{x\in\Xcal,\theta\in\Theta}\abr{\ddot{\mu}(\dotp{x}{\theta})}/\dot{\mu}(\dotp{x}{\theta})$.
\end{assumption}
Many widely studied GLB instances naturally satisfy these assumptions~\cite{sawarni25GLB}. For example, the triple $(\mu(z),\Lcal_\mu,M_\mu)$ takes the form $(z,1,0)$ in linear, $(\frac{1}{1 + e^{-z}},\frac{1}{4},1)$ in logistic, and $(e^z,e,1)$ in Poisson.

\subsection{Confidence bound for the true parameter}\label{sec:CB parameter}
An important step in analyzing the regret bound of the algorithm is to establish a confidence set for the underlying parameter $\theta^*$. 
This involves constructing a region that reliably contains $\theta^*$ throughout the learning process. 
To obtain a practical and tighter bound, we adopt confidence sets derived from the log-likelihood function using an \emph{ellipsoidal relaxation}. The confidence width is provided by recent work~\cite{lee2025unified}, which can be substituted with alternative, potentially tighter bounds.
\begin{lemma}[Adapted from Theorem 3.2. in \citet{lee2025unified}]\label{lem:beta}
Let $\Lcal_t:=\max_{\theta\in\Theta}\|\nabla L_t(\theta)\|$ denote the Lipschitz constant of  the loss function $L_t(\cdot)$.\footnote{
It has been shown by \citet{lee2025unified} that $\Lcal_t=\Ocal(t)$ for linear, logistic, and Poisson bandit instances.}
For any $\lambda>0$, define the regularized Hessian matrix at $\htt$ as $\hat{H}_t:=\nabla^2 L_t(\htt)+\lambda\Ib$. 
Then, with probability at least $1-\delta$, for all $t\geq1$, it holds that $\theta^*\in\Theta_t(\delta,\lambda):=\{\theta\in\RR^d\mid\|\theta-\htt\|_{\hat{H}_t}\leq\beta_t(\delta)\}$, where
\begin{equation*}
\beta_t(\delta) = \sqrt{4\lambda + 2(1 + M_\mu)\rbr{\log\frac{1}{\delta}+d\log\rbr{\frac{2e\Lcal_t}{d}}}}.
\end{equation*}
\end{lemma}

\subsection{Concentration and anti-concentration}\label{sec:conc anti}
In addition to ensuring that $\hat{\theta}_t$ remains close to the true parameter $\theta^*$, it is crucial to balance the degree of randomization in the algorithm. Our regret analysis relies on showing that, with an appropriate choice of the tuning parameter ${c_t}$, the perturbed feature $\tilde{x}_t$ is stochastically optimistic, while concentrated around its estimated value $\mu(x_t^\top\hat{\theta}_t)$. These properties are fundamental to the analysis, and we introduce the relevant components in this section. 

\begin{definition}\label{def:events}
    Let $t\in[T]$. We define the following events:
    \begin{enumerate}[leftmargin=*, label=(\roman*)]
    \item $\hat{E}_t$: $\hat{\theta}_\tau$ remains close to $\theta^*$ for all steps $\tau \leq t$.
    \item $\tilde{E}_t$: all perturbed vectors $\tilde{x}_{\tau i}$ concentrated around their corresponding $x_{\tau i}$ for all steps $\tau \leq t$.
    \end{enumerate}
    For a given confidence level $\delta\in(0,1)$, define $\delta'=\delta/(4T)$ and $\gamma_t(\delta)=\beta_t(\delta')\sqrt{c\log(c'/\delta)}$, where $c$ and $c'$ are constants consistent with Eq.~\eqref{eqn:concentration}. The events are formally defined as:
    \begin{equation*}
        \begin{aligned}
            \hat{E}_t:=\cbr{\forall\tau\leq t;~\|\htt-\theta^*\|_{\hat{H}_t}\leq\beta_t(\delta')}\quad\text{and}\quad
            \tilde{E}_t:=\cbr{\forall\tau\leq t,x_{\tau i}\in\Xcal_\tau;~\tx_{\tau i}\in\Ecal_{\tau}(x_{\tau i})},
        \end{aligned}
    \end{equation*}
    where $\Ecal_t(x):=\{\tx\in\RR^d \mid |\dotp{\tx-x}{\htt}|\leq\gamma_t(\delta')\|x\|_{\hat{H}_t^{-1}}\}$ represents a high-probability region for the perturbed feature vector associated with each arm $x$.\label{eq:ellipsoid of arm(GLM)}
\end{definition}
A key requirement for the perturbation distribution $\Dcal$ is the following concentration property: for some constants $c, c' > 0$ and any unit vector $u$, we have
\begin{equation}\label{eqn:concentration}
    \PP_{\zeta\sim\Dcal}\rbr{|u^\top\zeta|\leq\sqrt{c\log(c'/\delta)}}\geq1-\delta.
\end{equation}
\paragraph{Remark.} The perturbing distribution of TS on $\theta$ is described in \cref{app:FPTS}. Unlike in TS, the concentration here is evaluated along a fixed direction $u$, not over all coordinates. This avoids the need for a union bound over $d$ dimensions, and the resulting bound is independent of $d$. This dimensionality reduction is a key advantage of perturbing features instead of parameters.

By construction, the events satisfy the nested structure $\hat{E}_T \subset \cdots \subset \hat{E}_1$ and $\tilde{E}_T \subset \cdots \subset \tilde{E}_1$. Building on these definitions, we show that the proposed perturbation distribution induces an appropriate balance between exploration and exploitation, as formalized in the following lemmas.
\begin{lemma}[Concentration] \label{lem:conc_event}
    Under Assumptions \ref{asm:Boundedness} and \ref{asm:Linkfunction}, with $c_t=\beta_t(\delta')$, $\PP(\hat{E}_T\cap\tilde{E}_T)\geq1-\frac{\delta}{2}$.
\end{lemma}
\begin{lemma}[Stochastic optimism] \label{lem:anti_event}
For each round $t$, given that the events $\hat{E}_t$ and $\tilde{E}_t$ occur, the probability of anti-concentration, conditioned on the filtration $\Hcal_{t-1}$ and under Assumptions \ref{asm:Boundedness} and \ref{asm:Linkfunction}, is lower bounded as 
\begin{equation*}
    \PP_t(\mu(\tx_t^\top\htt)\geq\mu(x_{t*}^\top\theta^*)\mid\hat{E}_t,\tilde{E}_t)\geq\frac{1}{4\sqrt{e\pi}}.
\end{equation*}
\end{lemma}
In our construction, the concentration term in Eq.~\eqref{eqn:concentration} induces only a constant-order inflation, yielding both $\beta_t$ and $\gamma_t$ scaling as $\Ocal(\sqrt{d})$. In contrast, TS incurs an extra $\sqrt{d}$ factor, resulting in $\gamma_t = \Ocal(d)$. This leads to a tighter exploration term in our algorithm and improved regret performance.

\subsection{Regret bound of \texorpdfstring{$\glmfp$}{GLM-FP}}
The complexity of the GLB problem is fundamentally determined by the following quantities, which captures the degree of nonlinearity in the reward function:
\begin{align}
    \kappa_*:= \frac{\sum_{t=1}^T\dot{\mu}(x_{t*}^\top\theta^*)}{T},\quad
    \kappa:=\min_{x\in\Xcal_{[T]},\theta\in\Theta} \dot{\mu}(x^\top\theta),\quad\text{where}\quad\Xcal_{[T]}:=\bigcup_{t=1}^T \Xcal_t. \label{eqn:min derivative}
\end{align}
These may scale exponentially small, particularly in the case of logistic bandits~\cite{Faury2020Improveds}. The following theorem presents the regret guarantee for our algorithm.
\begin{theorem}\label{thm:regret}
    For all $\delta\in(0,1)$, define $\delta'=\delta/(4T)$. Under Assumptions \ref{asm:Boundedness} and \ref{asm:Linkfunction}, with $c_t=\beta_t(\delta')$ and $\lambda=\Ocal(d)$, the cumulative regret $R(T)$ is bounded with probability at least $1-\delta$ as follows:
    \begin{align*}
        R(T)=\otil\rbr{d\sqrt{\kappa_*T}+d^2/\kappa}.
    \end{align*}
\end{theorem}
\vskip -0.1in
\textbf{Discussion of \cref{thm:regret}.} 
The leading term of the regret guarantee is $\otil(d\sqrt{\kappa_* T})$, which matches the minimax optimal regret bound in terms of the dimensionality $d$, the horizon $T$, and the instance-dependent constant $\kappa_*$~\cite{Abeille2020InstanceWiseMA, lee2025unified}. 
While RandUCB~\cite{vaswani20old}, another randomized algorithm, also achieves a regret bound of $\otil(d\sqrt{T})$, it is penalized by its inverse dependence on $\kappa$, lacking adaptation to instance-dependent complexity.
In contrast, to the best of our knowledge, our result is the first to show that a randomized algorithm achieves a regret bound with linear $d$-dependency, without additional dependence on the number of arms, benefiting form $\kappa_*$ in GLB problems.

\subsection{Proof sketch of \texorpdfstring{\cref{thm:regret}}{Theorem 1}}\label{sec:proofsketch}
Our proof begins by decomposing the instantaneous regret into two components: $\Reg_\text{FP}$, the regret which arises from the randomization through FP, and $\Reg_\text{EST}$, the regret accounting for the estimation error. By bounding each component separately, we derive the overall regret bound:
\begin{equation}\label{eqn:regret}
    R(T) = \underbrace{\sum_{t=1}^T \Big( 
        \overbrace{\mu(x_{t*}^\top\theta^*) - \mu(\tilde{x}_t^\top\hat{\theta}_t)}^{\text{Optimism}} 
        + \overbrace{\mu(\tilde{x}_t^\top\hat{\theta}_t) - \mu(x_t^\top\hat{\theta}_t)}^{\text{Perturbation Concentration}} 
    \Big)}_{\Reg_\text{FP}} 
    + \underbrace{\sum_{t=1}^T \left( \mu(x_t^\top\hat{\theta}_t) - \mu(x_t^\top\theta^*) \right)}_{\Reg_\text{EST}}.
\end{equation}
To bound $\Reg_\text{FP}$, we rely on two key properties: (i) optimism induced by selecting the best estimated arm, and (ii) concentration of the perturbation around the original context. Under the high-probability events $\hat{E}_T$ and $\tilde{E}_T$, both properties are well controlled, leading to the following bound:
\begin{align*}
\Reg\text{FP} \leq (8\sqrt{e\pi}+1) \sum_{t=1}^T \left| \max_{x \in \Ecal_t(x_t)} \mu(x^\top\hat{\theta}_t) - \mu(x_t^\top\hat{\theta}_t) \right| + \otil\left( \sqrt{\frac{d^2 T}{\lambda^2}} \right),
\end{align*}
where the $\otil(\cdot)$ term results from an Azuma–Hoeffding concentration and becomes negligible under the choice $\lambda = \Ocal(d)$. Note that since the stochastic optimistic probability in \cref{lem:anti_event} is lower bounded by a constant, $\Reg_\text{FP}$ can be effectively bounded by the sum of per-round concentration widths, multiplied by a constant that is independent of $d$ and $T$.

Unlike previous randomized algorithms~\cite{abeille2017Linear, kveton2020randomized, vaswani20old} that linearize the reward function and thereby suffer regret bounds inversely proportional to $\kappa$, our analysis avoids such linearization. Instead, we directly utilize the gradient of the link function $\mu$ to characterize the shape of the elliptical confidence region, enabling more efficient exploration tailored to the reward model. This is reflected in the weighted Gram matrix $\hat{H}_t$, where the curvature of $\mu$ enters via the term $\dot{\mu}(x_\tau^\top\hat{\theta}_t)$.

However, this construction introduces a dependency on $t$ (rather than solely on $\tau$) in the weighted Gram matrix, which precludes a direct application of the standard \emph{Elliptical Potential Lemma} (EPL). To address this, we build upon recent analytical developments~\cite{Abeille2020InstanceWiseMA, Faury2020Improveds, lee2024improved}, and introduce a lower envelope of derivatives by defining $\bar{\theta}_t$ as the minimizer of $\dot{\mu}(x_t^\top\theta)$ over the union $\cup_{\tau \in [t,T]} \Theta_\tau(\delta, \lambda)$.

This yields a matrix $\bar{H}_t := \lambda \Ib + \sum_{\tau=1}^{t-1} \dot{\mu}(x_\tau^\top\bar{\theta}_\tau) x_\tau x_\tau^\top$ which satisfies $\hat{H}_t \succeq \bar{H}_t$, thereby enabling the application of EPL. We bound $\Reg_\text{EST}$ in a similar fashion and show that both $\Reg_\text{FP}$ and $\Reg_\text{EST}$ admit the same upper bound. This reduces the analysis to solving a quadratic inequality of the form:
\begin{equation*}
    \Reg_{\max} \leq A \sqrt{B + C \Reg_{\max}} + D,\quad\text{where}\quad \Reg_{\max} = \max\{\Reg_\text{FP}, \Reg_\text{EST}\}.
\end{equation*}
The constants $A$, $B$, $C$, and $D$ are explicitly analyzed in the full proof, deferred to \cref{app:main proof}.

\section{Carving off the \texorpdfstring{$\sqrt{d}$}{root d} factor compared to TS}\label{sec:discussion}
The proposed algorithm, $\glmfp$, adopts a novel exploration strategy by perturbing the input feature vectors, in contrast to conventional randomized algorithms such as Thompson Sampling (TS), which introduce randomness into the model parameter $\theta$. This design yields a regret bound with linear dependence on $d$, whereas TS incurs a higher $\Ocal(d^{3/2})$ dependence. We examine the origin of this discrepancy by comparing the linear variants of both algorithms (see \cref{app:linear}), highlighting how each introduces randomness to facilitate exploration.

The randomized evaluation score $\tilde{f}_t(x_i)$ (either $x_{ti}^\top\hat{\theta}_t$ for TS or $\tilde{x}_{ti}^\top\hat{\theta}_t$ for FP) for each arm used for action selection in both algorithms is straightforward to compute as follows:
\begin{equation*}
    \text{(TS)}\quad \tilde{f}_t(x_i)=x_{ti}^\top\tilde{\theta}_t=x_{ti}^\top\hat{\theta}_t+c_t\,x_{ti}^\top V_t^{-1/2}\zeta_t,\qquad
    \text{(FP)}\quad \tilde{f}_t(x_i)=\tilde{x}_{ti}^\top\hat{\theta}_t=x_{ti}^\top\hat{\theta}_t+c_t\,z_t\|x_{ti}\|_{V_t^{-1}},
\end{equation*}
where $\zeta_t\!\sim\!\Ncal(\zero,\Ib)$ and $z_t\!\sim\!\Ncal(0,1)$. Thus, for each arm individually, both methods induce the same Gaussian distribution,\footnote{In the linear bandit setting, this distribution also matches RandUCB~\citep{vaswani20old}, though its derivation is conceptually distinct and diverges beyond the GLB case; see \cref{sec:intuition,app:relations}.} $\tilde{f}_t(x_i)\!\sim\!\Ncal(x_{ti}^\top\hat{\theta}_t,\,c_t^2\|x_{ti}\|_{V_t^{-1}}^2)$.  
However, the \emph{object of perturbation}—parameter in TS versus feature in FP—fundamentally alters how exploration bonuses are assigned and how arm comparisons are coupled at each timestep.

\begin{figure*}[t!]
    \centering
    \begin{subfigure}[t]{0.65\textwidth}
        \centering
        \includegraphics[width=\linewidth]{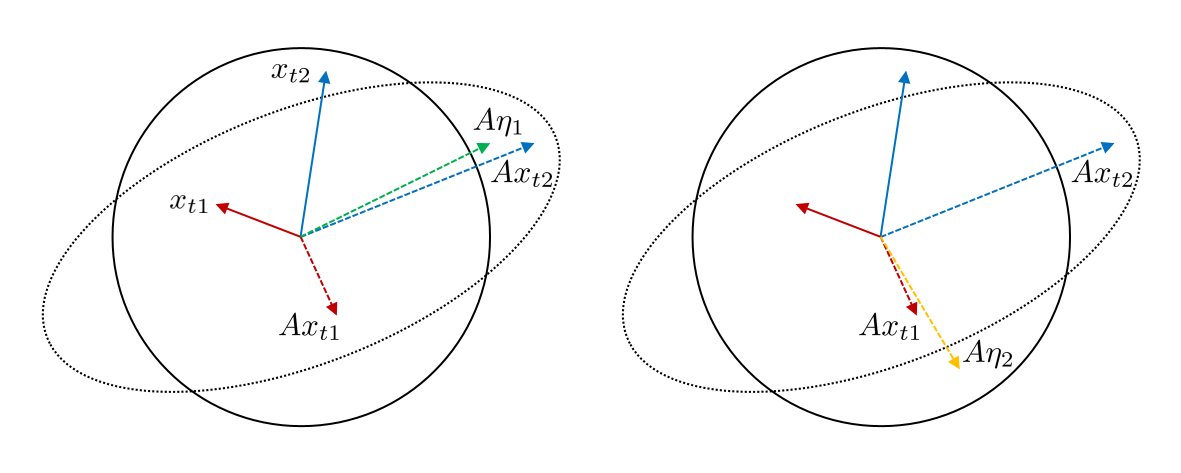}
        \vskip -0.1in
        \caption{Illustration of the transformation of the feature vectors in TS.}
        \label{fig:why}
    \end{subfigure}%
    ~ 
    \begin{subfigure}[t]{0.3\textwidth}
        \centering
        \includegraphics[width=\linewidth]{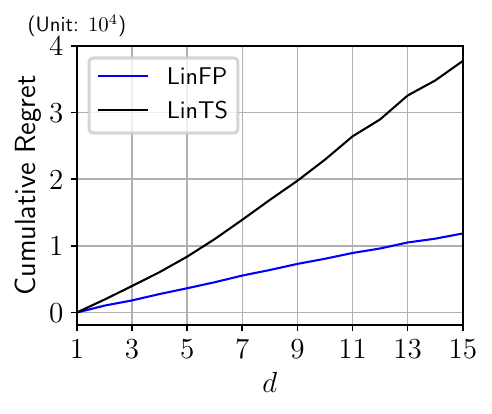}
        \vskip -0.1in
        \caption{Cumulative regret vs. $d$.}
        \label{fig:regret vs d}
    \end{subfigure}
    \caption{
    \textbf{(a)} Transformation of the well-explored arm $x_{t1}$ and under-explored arm $x_{t2}$ using $A = D^{1/2}P^\top$, where $V_t^{-1/2}=PDP^\top$. 
    Left: $\zeta_1$ induces a proper bonus. Right: $\zeta_2$ reverses the effect. 
    \textbf{(b)} Average terminal regret $R(T)$ over $100$ runs with $T = 200{,}000$, $K = 50$, and varying $d$.
    }
\vspace{-1em}
\end{figure*}

In TS, the bonus $\langle x_{ti},\zeta_t\rangle_{\smash{V_t}^{-1/2}}$ projects each arm onto a shared random direction $\zeta_t$ in the $V_t^{-1/2}$-transformed space. As conceptually illustrated in \cref{fig:why}, this shared dependence can produce counterintuitive effects: well-explored arms may occasionally receive large bonuses simply due to alignment with $\zeta_t$, while under-explored ones may be neglected. Because the same $\zeta_t$ governs all arms, the analysis must ensure uniform reliability of exploration across directions, requiring high-probability control of the $d$-dimensional Gaussian vector $\zeta_t$. Applying a union bound over $d$ coordinates introduces an additional $\sqrt{d}$ factor into the regret bound.  

FP, in contrast, decouples exploration from directional uncertainty. Its bonus $z_t\cdot\|x_{ti}\|_{V_t^{-1}}$ scales directly with per-arm uncertainty, ensuring under-explored arms systematically receive larger bonuses. Since randomness enters only through the scalar $z_t$, the analysis reduces to bounding a one-dimensional Gaussian projection $u^\top\zeta_t$ for some fixed unit vector $u$. 

From an equation-level viewpoint, the regret of TS also admits the decomposition in Eq.~\eqref{eqn:regret}~\citep{abeille2017Linear}. 
For both algorithms, the estimation component $\Reg_{\mathrm{EST}}$ is bounded in the same manner, and the anti-concentration event occurs with the same probability $p$. 
Consequently, the optimism-driven term in the regret scales with the concentration width divided by $p$. 
The essential difference therefore lies in the concentration width---or, equivalently, in how each algorithm controls the perturbation magnitude that also determines the second part of the decomposition. We express the upper bound for each perturbation term as
\begin{align*} \text{(TS)}\quad&\big|x_t^\top(\tilde\theta_t-\hat\theta_t)\big|=c_t\big|x_{ti}^\top V_t^{-1/2}\zeta_t\big|\leq c_t\|x_{ti}\|_{V_t^{-1}}\cdot\|V_t^{-1/2}\zeta_t\|_{V_t}=c_t\|x_{ti}\|_{V_t^{-1}}\cdot\big|\zeta_t\big|,\\ 
\text{(FP)}\quad&\big|(\tilde{x}_t-x_t)^\top\hat{\theta}_t\big|=c_t\big|\|x_t\|_{V_t^{-1}}\frac{\hat{\theta}_t^\top\zeta_t}{|\hat{\theta}_t|_2}\big|=c_t\|x_{ti}\|_{V_t^{-1}}\cdot\big|u^\top\zeta_t\big|. \end{align*}
The concentration width in TS depends on $\|\zeta_t\|_2$, the norm of a $d$-dimensional Gaussian vector, whereas that of FP scales with the one-dimensional projection $|u^\top\zeta_t|$. 
To obtain a uniform high-probability guarantee across all coordinates, a union bound over the $d$-dimensional perturbation space introduces an additional $\Ocal(\sqrt{d})$ factor for TS. 
Thus, despite having identical marginal distributions for individual arms, the two algorithms differ fundamentally in how their perturbations couple across arms: TS requires concentration over all directions in $\RR^d$, whereas FP relies on a single scalar randomization. 
This structural decoupling eliminates the extraneous $\sqrt{d}$ factor, yielding linear $\Ocal(d)$ dependence in the regret bound and clarifying the geometric origin of FP’s improvement over TS.

\section{Experiments} \label{sec:experiments}
We conduct experiments in two contextual bandit settings: (i) generalized linear bandits (GLBs), including linear and logistic models, and (ii) nonlinear contextual bandits based on neural networks. In each setting, we compare the proposed method with state-of-the-art baselines across varying feature dimensions and datasets. All results are averaged over $100$ independent runs to ensure robustness. Detailed experimental setups are provided in \cref{app:additional exp}.

\subsection{Generalized linear bandits} \label{sec:exp:glb}
We evaluate $\glmfp$ in both linear and logistic contextual bandit settings, where the expected reward follows a generalized linear model. 
In the linear bandit case, the reward is generated as $r_t = x_t^\top\theta^* + \varepsilon_t$ with $\varepsilon_t\sim\Ncal(0,1)$, while in the logistic bandit case, $r_t \sim \mathrm{Bernoulli}(\mu(x_t^\top\theta^*))$ with the logistic function $\mu$. We compare against widely used baselines including $\varepsilon$-greedy, UCB, TS, PHE, and RandUCB. Parameter estimation is performed via regularized weighted least squares (WLS) in linear bandit setting or IRLS in logistic bandit setting.

\begin{figure*}[t]
\vskip -0.1in
\begin{center}
\begin{subfigure}[t]{\textwidth}
    \centering
    \includegraphics[width=\textwidth]{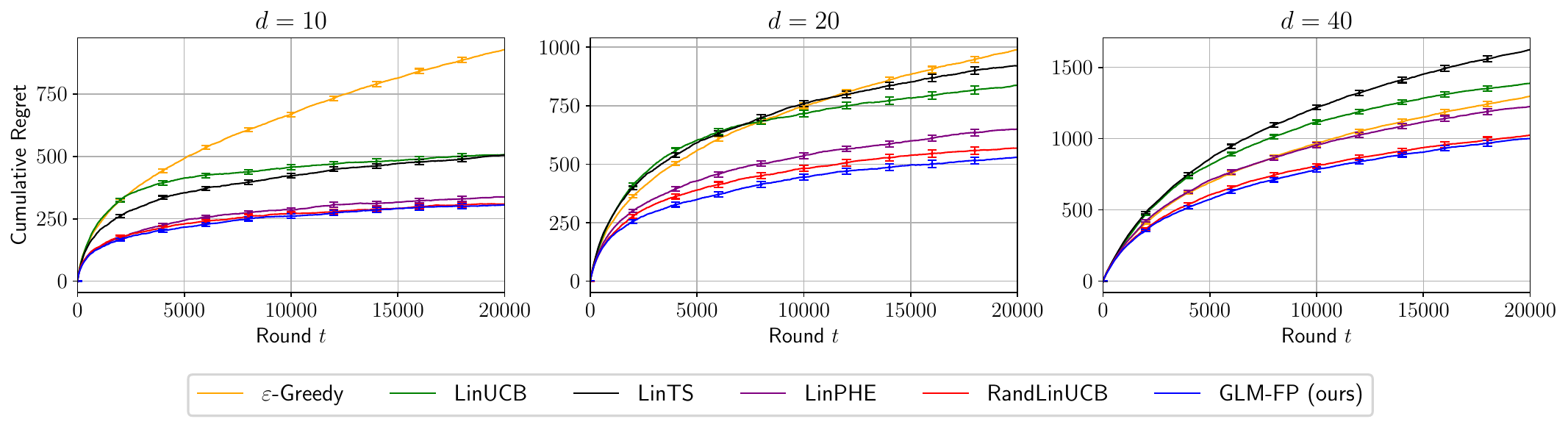}
    \label{fig:linear-bandits}
\end{subfigure}
\vskip -0.1in
\begin{subfigure}[t]{\textwidth}
    \centering
    \includegraphics[width=\textwidth]{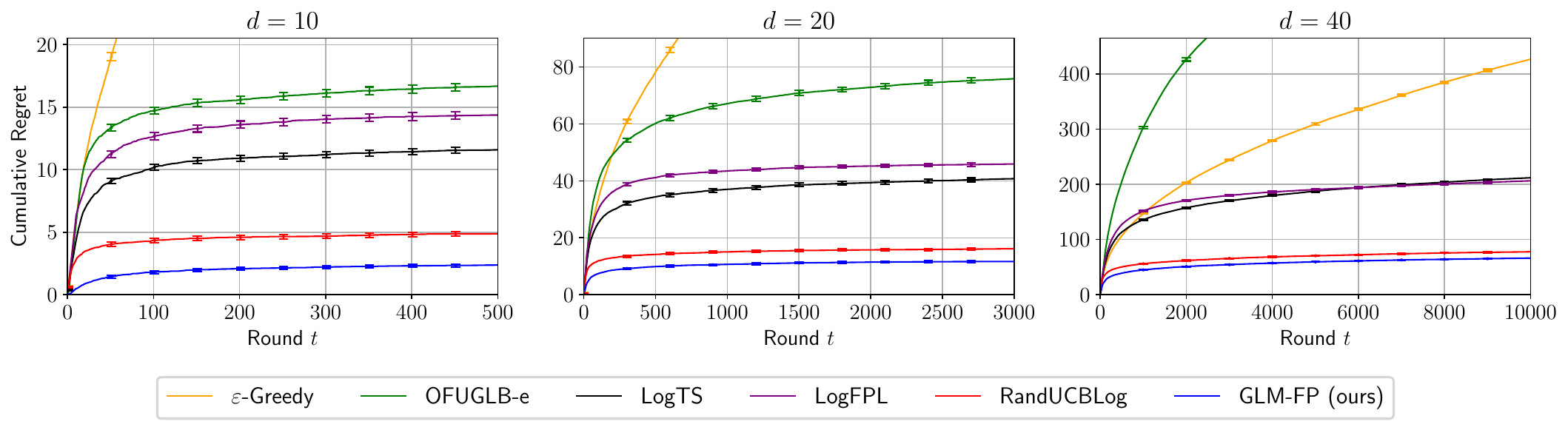}
    \label{fig:logistic-bandits}
\end{subfigure}
\vskip -0.1in
\begin{subfigure}[t]{\textwidth}
    \centering
    \includegraphics[width=\textwidth]{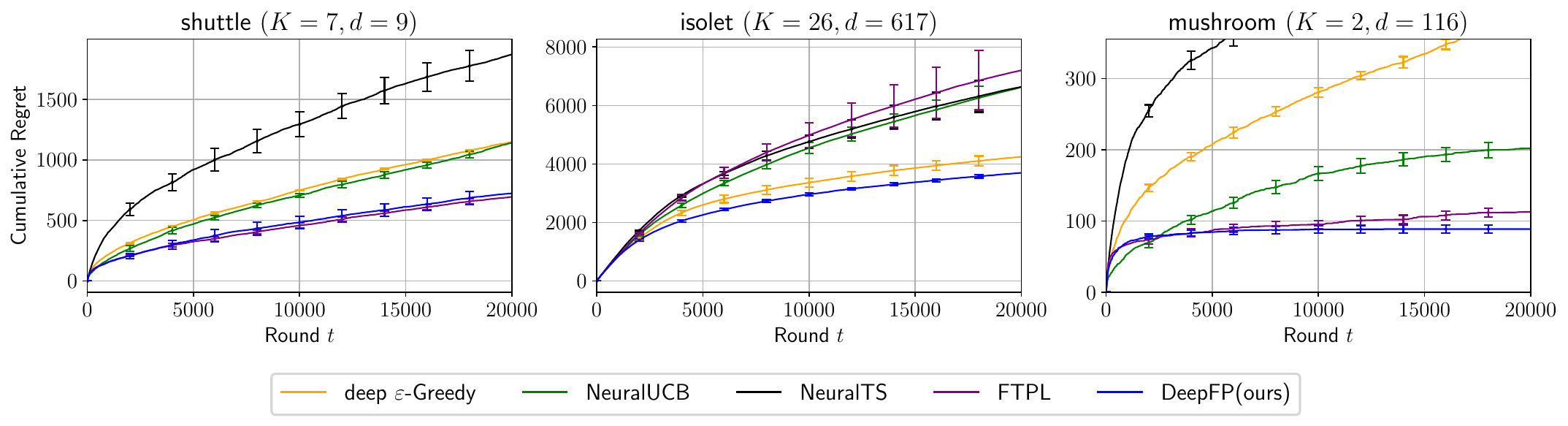}
    \label{fig:neural-bandits}
\end{subfigure}
\vskip -0.1in
\caption{Comparison of cumulative regret across contextual bandit algorithms: linear (top), logistic (middle), and neural (bottom).}
\label{fig:all-bandits}
\end{center}
\vskip -0.2in
\end{figure*}

As shown in \cref{fig:all-bandits} (top and middle), $\glmfp$ consistently achieves the lowest regret across all tested dimensions. While RandLinUCB performs competitively in the linear case, $\glmfp$ exhibits superior robustness, particularly in fixed-arm settings or environments with non-stationary arm sets (see \cref{app:additional exp} for more details). In the logistic setting, $\glmfp$ outperforms all baselines across all tested configurations, demonstrating its effectiveness even when the utility scale (i.e., $x^\top\theta^*$, the inner product term inside $\mu(\cdot)$) or the reward noise variance is varied. These results highlight the reliability and adaptability of our approach across diverse GLB scenarios.

\subsection{Neural bandits} \label{sec:exp:neural}
We extend the feature perturbation framework to the neural contextual bandit setting through a simple and scalable algorithm, $\deepfp$. At each round, $\deepfp$ injects independent, arm-wise Gaussian noise into the input features before prediction: $\tilde{x}_{ti} = x_{ti} + \zeta_{ti}$, where $\zeta_{ti} \sim \Ncal(0, I/t)$. The variance decay $1/t$ reflects the diminishing feature uncertainty over time, analogous to the confidence scaling $\|x\|_{\hat{H}_t^{-1}}$ in GLBs. This approach enables exploration without requiring access to model parameters or gradients. We evaluate $\deepfp$ on three UCI benchmark datasets: \texttt{shuttle} (7 classes, 9 features), \texttt{isolet} (26 classes, 617 features), and \texttt{mushroom} (binary, 112 features). Baselines include $\varepsilon$-greedy, NeuralUCB, NeuralTS, and Follow-the-Perturbed-Leader (FTPL). 

Unlike NeuralTS, which adds randomness to predicted rewards (essentially functioning as a randomized UCB variant), or FTPL, which perturbs historical rewards, $\deepfp$ directly perturbs the \emph{input features}. This structural distinction simplifies implementation and preserves the exploration intent in a more principled way. Furthermore, unlike parameter-sampling-based approaches whose application becomes increasingly unstable in high-capacity models where $p \gg d$, $\deepfp$ perturbs a much lower-dimensional space, making it scalable and numerically stable.

As shown in \cref{fig:all-bandits} (bottom), $\deepfp$ outperforms all baselines across the three datasets. Notably, it achieves strong performance without relying on posterior approximations or gradient-based confidence intervals, demonstrating that simple feature perturbation can remain effective even in complex, high-dimensional settings.
\section{Conclusion}\label{sec:conclusion}
We introduced a new paradigm for randomized exploration in contextual bandits. By shifting stochasticity to feature space, FP bridges the gap between randomized and optimistic methods while achieving optimal regret. 
This perspective offers a broadly applicable exploration principle and invites future work on leveraging structured randomness for efficient decision making.


\section*{Acknowledgements}
This work was supported by the National Research Foundation of Korea~(NRF) grant funded by the Korea government~(MSIT) (No.  RS-2022-NR071853 and RS-2023-00222663), by Institute of Information \& communications Technology Planning \& Evaluation~(IITP) grant funded by the Korea government~(MSIT) (No. RS-2025-02263754), and by AI-Bio Research Grant through Seoul National University.

\bibliographystyle{plainnat}
\bibliography{body/0_references}


\clearpage
\appendix
\crefalias{section}{appendix}
\crefalias{subsection}{appendix}
\crefalias{subsubsection}{appendix}
\setcounter{tocdepth}{2}    

\counterwithin{table}{section}
\counterwithin{figure}{section}
\counterwithin{equation}{section}
\counterwithin{algorithm}{section}
\counterwithin{theorem}{section}
\counterwithin{lemma}{section}
\counterwithin{corollary}{section}
\counterwithin{proposition}{section}
\counterwithin{assumption}{section}
\counterwithin{definition}{section}
\counterwithin{remark}{section}
\counterwithin{condition}{section}

\addcontentsline{toc}{section}{Appendix}
\part{Appendix}
\parttoc
\section{Further related wokrs}\label{app:relations}

\paragraph{Landscape of GLB Algorithms.}
Generalized linear bandit (GLB) algorithms can be broadly divided into \emph{deterministic OFU-type} and \emph{randomized exploration}-based methods. 
OFU approaches such as GLM-UCB~\citep{filippi2010parametric} and Logistic-UCB~\citep{jun2017scalable} achieve the tight $\otil(d\sqrt{T})$ or $\otil(d\sqrt{T}/\kappa)$ regret bound, and refinements further improve confidence construction~\citep{jun21a}, achieving $\otil(\sqrt{dT\log K})$ in the finite-$K$ arm setting. 
\citet{Abeille2020InstanceWiseMA} provided an instance-dependent analysis showing that the regret can benefit from the curvature constant $\kappa_*$, achieving $\otil(d\sqrt{\kappa_*T})$. 
Subsequent works~\citep{Lee2024Improvedlog,lee2025unified} relaxed the dependence on $S$ or improved computational efficiency~\citep{faury22a} while preserving the same order. 
Randomized methods such as Thompson Sampling (TS) and Perturbed History Exploration (PHE)~\citep{abeille2017Linear,agrawal2013thompson,kveton2020perturbed,kveton2020randomized}, as well as more recent algorithms like EVILL~\citep{janz24a} and RandUCB~\citep{vaswani20old}, typically achieve superior empirical performance but incur an additional $\sqrt{d}$ penalty in the worst case, yielding $\otil(d^{3/2}\sqrt{T})$ for infinite arms or $\otil(d\sqrt{T\log K}/\kappa)$ for finite arms. 
Our proposed feature perturbation (FP) departs from parameter- or reward-perturbation by randomizing the \emph{input features}, thereby closing this gap: as summarized in Table~\ref{tab:taxonomy}, FP is the first randomized algorithm for GLBs to provably achieve $\otil(d\sqrt{\kappa_*T}+d^2/\kappa)$ regret with no dependence on $K$, unifying the tight guarantees of OFU-type methods with the empirical robustness of randomized exploration.

\begin{table}[t]
\centering
\caption{Representative GLB algorithms: regret bounds and source of stochasticity.}
\label{tab:taxonomy}
\begin{center}
\renewcommand{\arraystretch}{1.0}
\begin{tabular}{llll}
\toprule
\textbf{Type} & \textbf{Algorithm} & \textbf{Regret Upper Bound} & \textbf{Stochasticity} \\
\midrule
\multirow[c]{3}{*}{Deterministic}
& GLM-UCB \cite{filippi2010parametric} & $\tilde{\mathcal{O}}(d\sqrt{T}/\kappa)$ & -- \\
& Logistic-UCB-2 \cite{jun2017scalable} & $\tilde{\mathcal{O}}(d\sqrt{T} + d^2/\kappa )$ & -- \\
& OFUGLB \cite{lee2025unified} & \blue{$\tilde{\mathcal{O}}(d\sqrt{\kappa_* T} + d^2/\kappa )$} & -- \\
\midrule
\multirow[c]{6}{*}{Randomized}
& LinTS \cite{abeille2017Linear} & $\tilde{\mathcal{O}}(d^{3/2}\sqrt{T}/\kappa)$ & Parameter ($\theta$) \\
& GLM-TSL \cite{kveton2020randomized} & $\tilde{\mathcal{O}}(d^{3/2}\sqrt{T}/\kappa)$ & Parameter ($\theta$) \\
& GLM-FPL \cite{kveton2020randomized} & $\tilde{\mathcal{O}}(d^{3/2}\sqrt{T}/\kappa)$ & Reward ($r$) \\
& RandUCB \cite{vaswani20old} & $\tilde{\mathcal{O}}(d\sqrt{T}/\kappa)$ & Linear utility ($x^\top \theta$) \\
& \textbf{GLM-FP (Ours)} & \blue{$\bm{\tilde{\mathcal{O}}\left(d\sqrt{\kappa_* T} + d^2/\kappa \right)}$} & \textbf{Feature vector ($\bm x$)} \\
\bottomrule
\end{tabular}
\end{center}
\end{table}

\paragraph{Comparison to RandUCB Algorithm.}
Our linear variant $\linfp$ (see \ref{alg:LinFP}) and RandUCB~\citep{vaswani20old} coincide in the linear bandit setting. Both sample randomized scores $\tilde{f}_t(x_i)\!\sim\!\Ncal(x_{ti}^\top\hat{\theta}_t,\,\beta_t^2\|x_{ti}\|_{\smash{V_t}^{-1}}^{2})$ and couple the arms identically, yielding matching $\tilde{\Ocal}(d\sqrt{T})$ regret bounds. The difference lies only in the \emph{source} of randomness: FP perturbs the input features, whereas RandUCB perturbs the reward estimate itself. These distinct mechanisms collapse to the same Gaussian rule under linear models, though they have been analyzed through different theoretical perspectives. Equivalently, one may view RandUCB as a special instance of FP with an identical perturbation distribution, differing only in interpretation and analytical framework.

In generalized linear bandits (GLBs), however, the two algorithms diverge fundamentally. RandUCB extends its linear recipe by linearizing the link function, introducing a multiplicative $\kappa^{-1}$ penalty and yielding a regret bound of $\tilde{\Ocal}(d\sqrt{T}/\kappa)$. In contrast, $\glmfp$ perturbs the inputs directly using a curvature-aware Gram matrix that weights past features by $\dot{\mu}(x^\top\hat{\theta}_t)$, enabling both anti-concentration (for exploration) and concentration (for confidence). This yields a tighter regret of $\tilde{\Ocal}(d\sqrt{\kappa_*T})$, linear in $d$ and directly in $\kappa_*$. Conceptually, FP injects stochasticity \emph{before} inference, so that each sampled reward $\hat{f}(\tilde{x})$ remains within the hypothesis class—preserving inductive bias and reflecting epistemic uncertainty. RandUCB, in contrast, adds randomness \emph{after} inference, producing post hoc scores that may not correspond to any $f\in\Fcal$. As a result, in expressive models FP remains aligned with the model structure, while RandUCB may misalign exploration incentives, leading to divergent empirical and theoretical behaviors.

\paragraph{Geometry and Scalability in Randomized Exploration.}
Recent studies have examined when randomized exploration can match the $\otil(d\sqrt{T})$ guarantees of optimistic approaches. \citet{whenandwhy} identified a class of geometric conditions—absorbing, strongly convex, and smooth action sets—under which Thompson Sampling (TS) achieves optimal dependence on $d$. While these conditions provide valuable insights into the role of geometry, they often fail to hold in high-dimensional or unstructured settings. Subsequent works~\citep{hamidi2020frequentist,kveton2020randomized} further clarified that posterior variance inflation can inherently introduce the extra $\sqrt{d}$ factor observed in randomized methods. In contrast, feature-level perturbation achieves similar statistical optimality under the standard boundedness assumption, bridging geometric optimality with more general feature-level regularity. From a computational standpoint, randomized exploration in large or continuous action spaces raises significant scalability challenges. Several strategies have been proposed to mitigate this issue, including lazy or delayed updates of the Gram matrix~\citep{abbasi2011improved}, two-stage candidate selection using approximate nearest neighbors, and optimization-oracle-based methods such as batched soft elimination~\citep{batchedlinear}. These approaches highlight a trade-off between statistical tightness and computational efficiency: while algorithms like FP prioritize theoretical optimality in the online setting, batched or oracle-based techniques offer scalable alternatives for large-scale practical applications.

\paragraph{Connections to Feature Perturbation in Broader ML}
Feature perturbation is a common idea in other areas of machine learning, most notably in computer vision and natural language processing, where it is employed for robustness or regularization during training. Examples include data augmentation~\citep{shortensurvey}, adversarial training~\citep{goodfellowexplaining}, or NoisyNets for exploration in deep reinforcement learning~\citep{noisynet}. In these contexts, perturbations are introduced at training time to improve model generalization or robustness. In contrast, our FP algorithm introduces perturbations at \emph{decision time} as a principled mechanism for exploration in online learning. This distinction highlights the novelty of FP: rather than making a static predictor robust, we leverage feature perturbations dynamically to induce stochasticity in action selection, enabling efficient exploration. The same principle applies naturally when contextual information comes from high-dimensional embeddings, such as ResNet or ViT features for images~\citep{he2016residual, dosovitskiy2021an} or BERT embeddings for language~\citep{devlin2019bert}, where FP can induce semantically meaningful exploration by perturbing compact representations. Thus, FP not only closes a theoretical gap in contextual bandits but also suggests a unifying exploration paradigm that resonates with broader trends in modern ML. Finally, this perspective also provides a bridge to reinforcement learning (RL), where perturbing the state–action feature representation can serve as an efficient alternative to parameter-space randomization used in posterior sampling or Noisy Networks~\citep{osband2013posterior, osband2018randomized, zanette2023frequentist}. Extending feature perturbation to structured settings such as Linear MDPs or value-function approximation is a promising direction for future work, potentially unifying exploration principles across bandit and reinforcement learning paradigms.

\section{Properties of FP distributions}\label{app:FPTS}
\paragraph{Perturbation in Thompson Sampling}
The perturbation distribution utilized in the TS algorithm to bring randomness to the parameter, as described by \citet{abeille2017Linear} is as followed:
\begin{definition}[Definition 1. in \citet{abeille2017Linear}]\label{def:D_TS}
    $\Dcal^{\text{TS}}$ is a multivariate distribution on $\RR^d$, absolutely continuous with respect to the Lebesgue measure, and satisfies the following properties:

    1. (Anti-concentration) There exists a positive probability $p>0$ such that for any unit vector $u\in\RR^d$,
    \begin{equation*}
        \PP_{\zeta\sim\Dcal^{\text{TS}}}(u^\top\zeta\geq1)\geq p,
    \end{equation*}
    2. (Concentration) There exist positive constants $c$ and $c'$ such that for all $\delta\in(0,1)$,
    \begin{equation*}
        \PP_{\zeta\sim\Dcal^{\text{TS}}}\rbr{\lnorm{\zeta}{}\leq\sqrt{cd\log(c'd/\delta)}}\geq1-\delta.
    \end{equation*}
\end{definition}

Below, we provide examples of distributions satisfying these \emph{anti-concentration} and \emph{concentration} properties, with the latter condition restated in Eq.~\eqref{eqn:concentration} in \cref{sec:analysis}.

\paragraph{Example 1: Gaussian distribution $\zeta\sim\Ncal(\zero,\Ib)$} The concentration property comes directly from \pref{lem:normal bound}, as the inner product of a standard multivariate normal random variable $\zeta$ and an arbitrary unit vector $u$ follows a standard normal distribution. In the same manner, for a unit vector $u$,
\begin{equation}\label{eqn:normal anti}
    \PP_{\zeta\sim\Ncal(\zero,\Ib)}\rbr{u^\top\zeta\geq1}=
    \PP_{z\sim\Ncal(0,1)}\rbr{z\geq1}=
    \frac{1}{2}\text{erfc}(\frac{1}{\sqrt{2}})\geq\frac{1}{4\sqrt{e\pi}}.
\end{equation}
Thus, the standard Gaussian distribution satisfies the concentration property with $c=c'=2$ and anti-concentration property with $p=\frac{1}{4\sqrt{e\pi}}$.
Adjusting the scale of the covariance matrix, we can easily prove other variants satisfy the conditions.

\paragraph{Example 2: Uniform distribution $\zeta\sim\Ucal_{B_d(\zero,\sqrt{d})}$} Let the random variable $\zeta=rv$, where $r=\norm{\zeta}\in[0,\sqrt{d}]$ and $v=\zeta/\norm{\zeta}$ is a unit vector. Then, $u^\top\zeta$ can be expressed as the product of two independent random variables, $r$ and $u^\top v$ ($\sim\text{Beta}(\frac{1}{2},\frac{d-1}{2})$), as $r\cdot(u^\top v)$. These two random variables follow the distributions:
\begin{align*}
    f_r(r)=\frac{dr^{d-1}}{\sqrt{d}^d},\quad r\in[0,\sqrt{d}],\quad\quad\text{and}\quad\quad
    f_{u^\top v}(x)=\frac{\Gamma(\frac{d}{2})}{\Gamma(\frac{1}{2})\Gamma(\frac{d-1}{2})}(1-x^2)^{\frac{d-3}{2}},\quad x\in[-1,1].
\end{align*}
Based on these random variables, we can write:
\begin{equation*}
    f_{u^\top\zeta}(z)=\int_0^{\sqrt{d}}\int_{-1}^1\delta(z-rx)f_r(r)f_{u^\top v}(x) dxdr.
\end{equation*}
Using Monte Carlo simulations, we observe that $u\top\zeta$ has a lighter tail distribution compared a standard normal distribution. Accordingly, $c=c'=2$ satisfies the concentration property. By proposition 9 and 10 in \citet{abeille2017Linear},
\begin{equation*}
    \PP(u^\top\zeta\geq1)=\frac{1}{2}I_{1-\frac{1}{d}}\rbr{\frac{d+1}{2},\frac{1}{2}}\geq\frac{1}{16\sqrt{3\pi}},
\end{equation*}
where $I_x(a,b)$ is the incomplete regularized beta function. This suggests that the Uniform distribution satisfies the anti-concentration property with $p=\frac{1}{16\sqrt{3\pi}}$.
\section{Application of FP algorithm}
\subsection{Generalized version of \texorpdfstring{$\glmfp$}{GLM-FP}}\label{app:general}
In \cref{sec:algorithm}, we introduced how FP can be applied to the contextual bandit settings in which the reward model extends beyond generalized linear models. We provide the general algorithmic framework below.
\begin{algorithm}[h!]
    \caption{Feature Perturbation in Bandit Problems}
    \begin{algorithmic}[1]
        \State {\bfseries Input:} Regularization parameter $\lambda > 0$, tuning parameter $\{c_t\}$
        \For{$t = 1, 2, \ldots, T$}
            \State Compute $\hat{f} = \argmin_{f \in \Fcal} \sum_{\tau = 1}^{t-1} (f(x_{\tau, i_\tau}) - r_\tau)^2$ via a least squares oracle
            \State Sample $\tilde{x}_{ti} \sim \Dcal(x_{ti}, \Sigma_{ti})$ for all $i$ 
            \Comment{e.g., $\Dcal(x_{ti}, \Sigma_{ti}) = \Ncal(x_{ti}, \Ib / t)$}
            \State Select arm $i_t = \arg\max_{i \in [|\Xcal_t|]} \hat{f}(\tilde{x}_{ti})$ \Comment{either i.i.d.\ or via shared perturbation}
            \State Observe reward $r_t = f^*(x_{t,i_t}) + \xi_t$
        \EndFor
    \end{algorithmic}
    \label{alg:FP}
\end{algorithm}
\subsection{Application to the linear bandit problem}\label{app:linear}
While line 4 in \cref{alg:FP} merely defines the sampling distribution for each arm, in practice—mirroring the design of $\glmfp$—one may introduce a shared perturbing factor that is first sampled and then applied to all arms. This construction induces dependencies across the perturbed arms and can serve as the basis for the arm selection mechanism.

\begin{algorithm}[h!]
    \caption{$\linfp$: Feature Perturbation in Linear bandits}
\begin{algorithmic}[1]
    \State {\bfseries Input:} Regularization parameter $\lambda>0$, tuning parameter $\{c_t\}$
    \State {\bfseries Initialize:} $V_1\gets \lambda \Ib$, $b_1\gets \zero_d$
    \For{$t=1,2,\ldots,T$}
        \State Compute $\htt=V_t^{-1}b_t$
        \State Sample $\zeta_t \sim \Ncal(\zero,\Ib)$\Comment{Shared perturbing factor}
        \State Compute $\tx_{ti}=x_{ti}+c_t \cdot\frac{\|x_{ti}\|_{\hat{V}_t^{-1}}}{\|\htt\|}\cdot \zeta_t$ for all $i$
        \State Choose $i_t=\arg\max_{i\in[|\Xcal_t|]}\tx_{ti}^\top\htt$ \Comment{$x_{ti}^\top\tilde{\theta}_t\sim \Ncal(x_{ti}^\top\hat{\theta}_t, c_t^2\|x_{ti}\|_{\smash{V_t^{-1}}}^2)$ for all $i$.}
        \State Observe reward $r_t=x_{t,i_t}^\top\theta^*+\xi_t$
        \State Update$V_{t+1}=V_t+x_{t,i_t}x_{t,i_t}^\top,\quad b_{t+1}=b_t+x_{t,i_t}r_t$
    \EndFor
\end{algorithmic}
\label{alg:LinFP}
\end{algorithm}

In \cref{sec:discussion}, we compare $\linfp$, the linear variant of our approach, to LinTS~\cite{abeille2017Linear, agrawal2013thompson}. The primary algorithmic difference lies in lines 6–7 of \cref{alg:LinFP}. In our method, the shared random vector $\zeta_t$ is used to perturb each feature vector. In contrast, LinTS perturbs the model parameter as $\tilde{\theta}_t = \hat{\theta}_t + c_t \cdot V_t^{-1/2} \zeta_t$, and selects the arm maximizing $x_{ti}^\top \tilde{\theta}_t$. For LinTS, we have:
\begin{align*}
    \EE[x_{ti}^\top \tilde{\theta}_t] &= x_{ti}^\top \hat{\theta}_t + c_t \cdot x_{ti}^\top V_t^{-1/2} \EE[\zeta_t] = x_{ti}^\top \hat{\theta}_t \\
    \Var[x_{ti}^\top \tilde{\theta}_t] &= c_t^2 \cdot \Var[x_{ti}^\top V_t^{-1/2} \zeta_t] = c_t^2 \cdot x_{ti}^\top V_t^{-1} x_{ti} = c_t^2 \|x_{ti}\|_{V_t^{-1}}^2
\end{align*}
While the marginal distribution of $x_{ti}^\top \tilde{\theta}_t$ under both methods is identical, the use of a shared perturbation $\zeta_t$ in $\linfp$ induces algorithmic coupling across arms. This distinction is further discussed in \cref{sec:discussion}.

\newpage
\section{Table of notations}
This section introduces additional notations and concepts essential for the analysis. For ease of reference, \cref{tab:notation} summarizes the key notations defined in this paper, along with newly introduced notations. Conventional concepts such as $d$, $T$, $\Acal$, $\Ccal$, $\Xcal$, $K$ or $r$ are omitted here. The concepts will be reintroduced as needed in subsequent sections.
\begin{table}[ht]
\caption{Notations and concepts for the analysis of the main theorem}
\label{tab:notation}
\begin{center}
\begin{adjustbox}{max width=\textwidth}
\renewcommand{\arraystretch}{1.5}
\begin{tabular}{llll}
\toprule
\textbf{Notation} & \textbf{Definition} \\
\midrule
$M_\mu$ & Self-concordance constant \\
$\Lcal_\mu$ & Lipschitz constant of the link function $\mu$ \\
$\Lcal_t$ & Lipshitz constant of the negative log-likelihood function \\
$\Theta_t(\delta,\lambda)$ & $1-\delta$ probability ellipsoidal relaxed confidence set with regularization $\lambda$ for the true parameter $\theta^*$ \\
$\beta_t(\delta)$ & $\sqrt{4S^2\lambda+2(1+SM_\mu)(\log(1/\delta)+d\log(2e\Lcal_t/d))}=\otil(\sqrt{d})$ \\
$\gamma_t(\delta)$ & $\beta_t(\delta/(4T))\sqrt{c\log(4c'T/\delta)}$ ($c,c'$: constant satisfying concentration property)\\
$\Ecal_t(x)$ & $\cbr{\tx\in\RR^d\big|\norm{\dotp{\tx-x}{\htt}}\leq\gamma_t(\delta/(4T))\norm{x}_{\hat{H}_t^{-1}}}$\\
$\kappa_*$ & Average derivative of link function at the true optimal arm over $T$ rounds \\
$\kappa$ & Minimum reachable derivative of link function \\ 
\midrule
\multicolumn{2}{l}{Warm-up stage} \\
\midrule
$I_T$ & $\cbr{t\in[T]:\rbr{\bnorm{\sqrt{\dot{\mu}(x_t^\top\btt)}x_t}_{\bar{H}_t^{-1}}\geq 1}\vee \rbr{\norm{x_t}_{V_t^{-1}}\geq1}}$ \\
\midrule
\multicolumn{2}{l}{Taylor remainder term} \\
\midrule
$\bar{\alpha}_t(x)$ & $\int_0^1(1-u)\dot{\mu}(x_t^\top\htt+u(x^\top\htt-x_t^\top\htt))du$ \\
$\bar{\alpha}_t(\theta,\nu)$ & $\int_0^1(1-u)\dot{\mu}\rbr{x_t^\top\theta+u(x_t^\top\nu-x_t^\top\theta)}du$ \\
\midrule
\multicolumn{2}{l}{Matrices} \\
\midrule
$V_t$ & $\sum_{\tau=1}^{t-1}x_\tau x_\tau^\top+\lambda\Ib$ \\
$\bar{V}_t$ & $\sum_{\tau=1}^{t-1}x_\tau x_\tau^\top+\lambda\Ib/\kappa$ \\
$\hat{H}_t$ & $\sum_{\tau=1}^{t-1}\dot{\mu}(x_\tau^\top\htt)x_\tau x_\tau^\top+\lambda\Ib$ \\
$\bar{H}_t$ & $\sum_{\tau=1}^{t-1}\dot{\mu}(x_\tau^\top\bar{\theta}_\tau)x_\tau x_\tau^\top+\lambda\Ib$ \\
$\tilde{H}_t(\theta,\nu)$ & $\sum_{\tau=1}^{t-1}\bar{\alpha}_\tau(\theta,\nu)x_\tau x_\tau^\top$ \\
\midrule
\multicolumn{2}{l}{Other notations} \\
\midrule
$R^*$ & $\max_{x\in\Xcal}|\mu(x^\top\theta^*)|$ \\
$\btt$ & $\argmin_{\theta\in\cup_{\tau\in[t,T]}\Theta_\tau(\delta,\lambda)}\dot{\mu}(x_t^\top\theta)$ \\
$(\tau(t),\omega_t)$ & $\argmax_{\tau\in[t,T],\theta\in\Theta_\tau(\delta,\lambda)}\abr{\mu(x_t^\top\theta)-\mu(x_t^\top\hat{\theta}_\tau)}$ \\ 
($p,c_{\delta'}$)& constants related to standard normal distribution ($p=1/(4\sqrt{e\pi})$, $c_{\delta'}=\sqrt{2\log(2/\delta')}$)\\
\bottomrule
\end{tabular}
\end{adjustbox}
\end{center}
\vskip -0.1in
\end{table}
\newpage
\section{Proof of main theorem}\label{app:main proof}
The first step in our proof is to derive the high-probability confidence bound for the estimate $\htt$. Using this bound, we ensure that the events defined in \cref{def:events} occur with high probability. Finally, we compute the regret bound of our algorithm under these events. 

\subsection{Proof of \texorpdfstring{\cref{lem:beta}}{Lemma 1}}\label{sec:proof gamma}
We rederived Theorem 3.2 of \citet{lee2025unified} to obtain a tighter bound with improved dependence on the regularization parameter $\lambda$, such that the $\lambda$ term no longer scales with $M_\mu$.

By Theorem 3.1. in \cite{lee2025unified}, with probability at least $1-\delta$, for all $t\geq1$, the following inequality holds:
\begin{align*}
    L_t(\theta^*)-L_t(\htt)\leq\log\frac{1}{\delta}+d\log\rbr{\frac{2eS\Lcal_t}{d}}:=\Wcal_t(\delta)^2
\end{align*}
Then we observe:
\begin{align}
    \int_0^1(1-u)\nabla^2L_t(\htt+&u(\theta^*-\htt)) du=\int_0^1(1-u)\sum_{\tau=1}^{t-1}\dot{\mu}(x_\tau^\top(\htt+u(\theta^*-\htt))x_\tau x_\tau^\top du\notag\\
    &=\sum_{\tau=1}^{t-1}\underbrace{\rbr{\int_0^1(1-u)\dot{\mu}(x_\tau^\top(\htt+u(\theta^*-\htt))du}}_{\bar{\alpha}_\tau(\htt,\theta^*)}x_\tau x_\tau^\top\notag=\tilde{H}_t(\htt,\theta^*),\label{eqn:tilde H}
\end{align}
where the second equality follows from Fubini's theorem, where the order of the integral and the summation can be switched. Using Taylor's theorem with an integral remainder (\cref{prop:Taylor}), we can further deduce that, with probability $1-\delta$:
\begin{align*}
    \Wcal_t(\delta)^2&\geq L_t(\theta^*)-L_t(\htt)\\
    &=\dotp{\nabla L_t(\htt)}{\theta^*-\htt}+(\theta^*-\htt)^\top\rbr{\int_0^1(1-u)\nabla^2L_t(\htt+u(\theta^*-\htt)) du}(\theta^*-\htt).
\end{align*}
As the optimality condition at $\htt$ infers that $\dotp{\nabla L_t(\htt)}{\theta^*-\htt}\geq0$ and by equation \cref{eqn:tilde H}, we have that
\begin{align*}
    \Wcal_t(\delta)^2&\geq\bnorm{\theta^*-\htt}_{\tilde{H}_t(\htt,\theta^*)}^2\geq\frac{1}{2+2SM_\mu}\bnorm{\theta^*-\htt}_{\hat{H}_t-\lambda\Ib}^2, \tag{\cref{prop:tildeH to nabla}}
\end{align*}
where the last inequality holds, since 
\begin{equation*}
    \tilde{H}_t(\htt,\theta^*)\succeq\sum_{\tau=1}^{t-1}\rbr{\frac{\dot{\mu}(x_\tau^\top\htt)}{2+2SM_\mu}}x_\tau x_\tau^\top=\frac{1}{2+2SM\mu}(\hat{H}_t-\lambda\Ib).
\end{equation*}
Accordingly,
\begin{align*}
    \bnorm{\theta^*-\htt}_{\hat{H}_t}^2\leq\bnorm{\theta^*-\htt}_{\hat{H}_t-\lambda\Ib}^2+\lambda\bnorm{\theta^*-\htt}^2\leq4S^2\lambda+2(1+SM_\mu)\Wcal_t(\delta)^2=\beta_t(\delta)^2,
\end{align*}
and by \cref{asm:Boundedness}, we finish a proof.
\subsection{Proof of Lemma \texorpdfstring{\ref{lem:anti_event}}{Lemma 2}}\label{proof anti}
Let the event $\ddot{E}_t$ be defined as $\ddot{E}_t := \cbr{\mu(\tx_t^\top\htt) \geq \mu(x_{t*}^\top\theta^*)}$. This event corresponds to the case where the chosen perturbed feature vector yields an optimistic expected reward relative to the true optimal arm at step $t$. To bound the anti-concentration probability as required in \cref{lem:anti_event}, we aim to lower bound $\PP_t(\ddot{E}_t \mid \hat{E}_t, \tilde{E}_t)$, conditioned on the events $\hat{E}_t$ and $\tilde{E}_t$. For any $t \in [T]$, we have:
\begin{align*}
    \PP_t\rbr{\ddot{E}_t\mid \hat{E}_t,\tilde{E}_t}=&\quad\PP_t\rbr{\mu(\tx_t^\top\htt)\geq \mu(x_{t*}^\top\theta^*)\mid\hat{E}_t,\tilde{E}_t}\\
    =&\quad\PP_t\rbr{\tx_t^\top\htt\geq x_{t*}^\top\theta^*\mid\hat{E}_t,\tilde{E}_t}\tag{$\because\mu$ is strictly increasing}\\
    \geq&\quad\PP_t\rbr{\tx_{t*}^\top\htt-x_{t*}^\top\htt\geq x_{t*}^\top\theta^*-x_{t*}^\top\htt\mid\hat{E}_t,\tilde{E}_t} \tag{$\because x_t=\underset{i\in[|\Xcal_t|]}{\argmax}~x_{ti}^\top\htt$}\\
    \geq&\quad\PP_t\rbr{\dotp{\tx_{t*}-x_{t*}}{\htt}\geq\abr{\dotp{x_{t*}}{\theta^*-\htt}}~\bigg|~\hat{E}_t,\tilde{E}_t}\\
    \geq&\quad\PP_t\rbr{\rbr{\beta_t(\delta')\frac{\norm{x_{t*}}_{\hat{H}_t^{-1}}}{\norm{\htt}}\zeta_t}^\top\htt\geq \norm{x_{t*}}_{\hat{H}_t^{-1}}\norm{\htt-\theta^*}_{\hat{H}_t}~\bigg|~\hat{E}_t,\tilde{E}_t}\\
    \geq&\quad\PP\rbr{\beta_t(\delta')\norm{x_{t*}}_{\hat{H}_t^{-1}}\cdot \dotp{\zeta_t}{u_t}\geq\beta_t(\delta')\norm{x_{t*}}_{\hat{H}_t^{-1}}\mid\hat{E}_t,\tilde{E}_t}\\
    =&\quad\PP\rbr{\dotp{\zeta_t}{u_t}\geq1}\geq\frac{1}{4\sqrt{e\pi}}:=p,
\end{align*}
where the third inequality follows from the Cauchy-Schwarz inequality, and the fourth from the assumption that under the event $\hat{E}_t$, we have $\norm{\theta^* - \htt}_{\hat{H}_t} \leq \beta_t(\delta')$. The final inequality follows from the anti-concentration property of the standard normal distribution, as detailed in \cref{eqn:normal anti}. For simplicity, we henceforth fix $p := 1/(4\sqrt{e\pi})$ as the corresponding lower bound on this probability.

\subsection{Proof of \texorpdfstring{\cref{lem:conc_event}}{Lemma 3}}\label{proof event}
We now proceed to establish how the probability of each event defined in \cref{def:events} can be ensured using the confidence bound $\beta_t(\delta)$ derived in \cref{sec:proof gamma}. Each event is analyzed and bounded individually, and the results are then combined to complete the proof of the lemma.
\paragraph{Bounding $\hat{E}_T$} Let $\delta'=\delta/(4T)$. By the choice of $\beta_t(\delta)$ in \cref{lem:beta}, we have that
\begin{align*}
    \forall1\leq t\leq T,&\quad \PP\rbr{\bnorm{\htt-\theta^*}_{\hat{H}_t}\leq\beta_t(\delta')}\geq1-\delta'\\
    \text{from union bound,}&\quad\PP\rbr{\bigcap_{t=1}^T\cbr{\bnorm{\htt-\theta^*}_{\hat{H}_t}\leq\beta_t(\delta')}}\geq1-\sum_{t=1}^T \PP\rbr{\bnorm{\htt-\theta^*}_{\hat{H}_t}\geq\beta_t(\delta')}\\
    \Longrightarrow&\quad\PP\rbr{\bigcap_{t=1}^T\cbr{\bnorm{\htt-\theta^*}_{\hat{H}_t}\leq\beta_t(\delta')}}\geq1-\sum_{t=1}^T\delta'\\
    \Longrightarrow&\quad\PP(\hat{E}_T)\geq1-T\delta'=1-\frac{\delta}{4}.
\end{align*}

\paragraph{Bounding $\tilde{E}_T$} The expression for the perturbed feature vector $\tx_{ti}$ is given as the expression $\tx_{ti}=x_{ti}+\beta_t(\delta')\frac{\norm{x_{ti}}_{\hat{H}_t^{-1}}}{\norm{\htt}}\zeta_t$ with the choice of $c_t=\beta_t(\delta')$, where $\zeta_t$ is drawn $\iid$ from $\Ncal(\zero,\Ib)$. Note that the constants $c$ and $c'$ indicated in \cref{eqn:concentration} are both $2$, as proved in \cref{eqn:normal anti}, from now on for simplicity, we let $c_{\delta'}:=\sqrt{2\log(2/\delta')}$. Since all arms are coupled\footnote{Uncoupled sampling means $\zeta_{ti}$'s are sampled for each arm respectively and it results in extra $\log K$ term in the regret bound because of the union bound over the number of arms at each step.} with same $\zeta_t$, we can write
\phantomsection\label{par:conc(linear)}
\begin{align}
    \forall 1\leq t\leq T,&\quad\PP\rbr{\forall x_{ti}\in\Xcal_t;\quad\abr{\dotp{\tx_{ti}-x_{ti}}{\htt}}\leq \gamma_t(\delta')\norm{x_{ti}}_{\hat{H}_t^{-1}}} \label{eqn:conc1}\\
    =&\quad\PP\rbr{\forall x_{ti}\in\Xcal_t;\quad\beta_t(\delta')\norm{x_{ti}}_{\hat{H}_t^{-1}}\abr{\dotp{\zeta_t}{\frac{\htt}{\norm{\htt}}}}\leq c_{\delta'}\cdot\beta_t(\delta')\norm{x_{ti}}_{\hat{H}_t^{-1}}}\label{eqn:conc2}\\
    =&\quad\PP\rbr{\abr{\dotp{\zeta_t}{u_t}}\leq c_{\delta'}}\geq1-\delta',\label{eqn:conc3}
\end{align}
where $u_t$ is a unit vector. The first equality holds by the definition of $c_{\delta'}$ and \cref{def:events}, and the inequality follows from the concentration property. The cancellation in the second equality plays a critical role in removing arm-wise dependence in GLB setting. A union bound over $T$ rounds yields
\begin{equation*}
    \PP(\tilde{E}_T)\geq1-T\delta'=1-\frac{\delta}{4}.
\end{equation*}
Finally, applying the union bound across the events $\hat{E}_T$ and $\tilde{E}_T$, we have that
\begin{equation*}
    \PP(\hat{E}_T\cap\tilde{E}_T)\geq1-\frac{\delta}{2}.
\end{equation*}

\textbf{Remark.} To guarantee the same probability level $1 - \delta'$ as in equations \cref{eqn:conc1}--\cref{eqn:conc3}, which bound the randomness arising from perturbations to the feature vectors, Thompson Sampling requires the confidence parameters $\beta$ and $\gamma$ to be inflated by an additional factor of $\sqrt{d}$. This inflation arises due to the right-hand side of the concentration bound in \cref{def:D_TS}, which scales with $\sqrt{d}$. Such adjustment is necessary to control the deviation in the perturbed estimated expected reward, which takes the form $x^\top(\tilde{\theta} - \hat{\theta})$. This is consistent with the reasoning discussed in \cref{sec:discussion}.

\subsection{Proof of \texorpdfstring{\cref{thm:regret}}{Theorem 1}}
In this section, we establish the regret guarantee for our algorithm. Given the complexity of the analysis, we divide the proof into multiple steps. Supporting lemmas and their proofs are deferred to \cref{app:supporting lemma}.

\paragraph{Step 1 (Warm-up)} We begin by partitioning the $T$ rounds into a ``warm-up'' stage and the primary stage. The set of time steps corresponding to the primary stage is defined as:
\begin{equation*}
    I_T:=\cbr{t\in[T]:\rbr{\bignorm{\sqrt{\dot{\mu}(x_t^\top\btt)}x_t}_{\bar{H}_t^{-1}}\leq 1}\wedge \rbr{\norm{x_t}_{\bar{V}_t^{-1}}\leq1}},
\end{equation*}
where $\bar{H}_t, \bar{V}_t$, and $\btt$ are given by:
\begin{align*}
    &\bar{H}_t:=\lambda\Ib+\sum_{\tau=1}^{t-1}\dot{\mu}(x_\tau^\top\bar{\theta}_\tau)x_\tau x_\tau^\top,\quad\quad \bar{V}_t:=\lambda\Ib/\kappa+\sum_{\tau=1}^{t-1}x_\tau x_\tau^\top,\quad\quad\btt:=\argmin_{\theta\in\cup_{\tau\in[t,T]}\Theta_\tau(\delta,\lambda)}\dot{\mu}(x_t^\top\theta).
\end{align*}
The introduction of $\bar{H}_t$ is crucial because $\hat{H}_t=\lambda\Ib+\sum_{\tau=1}^{t-1}\dot{\mu}(x_\tau^\top\htt)x_\tau x_\tau^\top$ depends on $t$, which prevents direct application of the Elliptical Potential Lemma (EPL; \cref{lem:EPL}), as discussed in \cref{sec:analysis}. To address this, we leverage $\bar{H}_t$, which incorporates the minimum derivative of $\mu$ within future confidence sets ($\bar{\theta}_\tau$), ensuring it serves as a smaller Gram matrix suitable for bounding the regret. Similarly, $\bar{V}_t$ is introduced to directly apply EPL.

Next, we bound each weighted $2$-norm using the Elliptical Potential Count Lemma (EPCL; \cref{lem:EPCL}), which guarantees that the regret incurred during the warm-up phase remains manageable. Consequently, the cumulative regret over $T$ rounds is decomposed as follows:
\begin{align*}
    R(T) &= \underbrace{\sum_{t \in I_T} \cbr{\mu(x_{t*}^\top\theta^*)-\mu(x_t^\top\theta^*)}}_{\Reg(T)} + \underbrace{\sum_{t \not\in I_T} \cbr{\mu(x_{t*}^\top\theta^*)-\mu(x_t^\top\theta^*)}}_{\text{warm-up regret}}\\
    &\leq \Reg(T) + 2R^*\rbr{\sum_{t=1}^T \ind{\bignorm{\sqrt{\dot{\mu}(x_t^\top\btt)}x_t}_{\bar{H}_t^{-1}}\geq 1}
    + \sum_{t=1}^T \ind{\norm{x_t}_{\bar{V}_t^{-1}}\geq1}} \\
    &\leq \Reg(T) + \frac{4d R^*}{\log 2}\cbr{\log\rbr{1+\frac{\Lcal_\mu}{\lambda\log2}}+\log\rbr{1+\frac{\kappa}{\lambda\log2}}},
\end{align*}
where $R^*:=\max_{x\in\Xcal}\abr{\mu(x^\top\theta^*)}$ is the maximum expected reward achievable under the underlying model.The first and the second inequalities hold from the definition of $I_T$ and by EPCL (Lemma~\ref{lem:EPCL}), respectively. Note that the warm-up regret is $\otil(d)$, which is independent of $T$.

\paragraph{Step 2-1 (Decomposition)} We decompose the cumulative regret for the primary stage into three components:
\begin{align*}
    \Reg(T) = \sum_{t \in I_T}\bigg(\underbrace{\cbr{\mu(x_{t*}^\top\theta^*)-\mu(\tx_t^\top\htt)}}_{A_t} + \underbrace{\cbr{\mu(\tx_t^\top\htt)-\mu(x_t^\top\htt)}}_{B_t} + \underbrace{\cbr{\mu(x_t^\top\htt)-\mu(x_t^\top\theta^*)}}_{C_t}\bigg).
\end{align*}
Here, $A_t$ and $B_t$ relate to the perturbations' effect on the estimated reward, while $C_t$ concerns the closeness of $\htt$ to $\theta^*$. We will bound each term under the events $\hat{E}_t$ and $\tilde{E}_t$.

\subparagraph{Bounding $C_t$} Bounding $C_t$ is straightforward. Using the confidence set $\Theta_t(\delta,\lambda)$, abbreviated as $\Theta_t$, we define $(\tau(t),\omega_t)$ as the pair maximizing the confidence width computed on the selected action at round $t$, $x_t$, after round $t$: $\argmax_{\tau\in[t,T],\theta\in\Theta_\tau}\abr{\mu(x_t^\top\theta)-\mu(x_t^\top\hat{\theta}_\tau)}$. Under the event $\hat{E}_t$, we know that $\theta^*\in\Theta_\tau$ for all $t\leq\tau\leq T$. Thus,
\begin{align*}
    C_t\ind{\hat{E}_t\cap\tilde{E}_t}
    =\rbr{\mu(x_t^\top\htt)-\mu(x_t^\top\theta^*)}\ind{\hat{E}_t\cap\tilde{E}_t}\leq
    \abr{\mu(x_t^\top\omega_t)-\mu(x_t^\top\htbt)}\ind{\hat{E}_t\cap\tilde{E}_t}.
\end{align*}

\subparagraph{Bounding $B_t$}  Let $\tx_t^*:=\argmax_{x\in\Ecal_t(x_t)}\abr{\mu(x^\top\htt)-\mu(x_t^\top\htt)}$. Under the event $\tilde{E}_t$, $\tx_t\in\Ecal_t(x_t)$ holds and we can write:
\begin{align*}
    B_t\ind{\hat{E}_t\cap\tilde{E}_t}
    =\rbr{\mu(\tx_t^\top\htt)-\mu(x_t^\top\htt)}\ind{\hat{E}_t\cap\tilde{E}_t}\leq\abr{\mu(\tx_t^{*\top}\htt)-\mu(x_t^\top\htt)}\ind{\hat{E}_t\cap\tilde{E}_t}.
\end{align*}
Before proceeding, note that for $\tx_t\in\Ecal_t(x_t)$, we can derive an upper bound using Taylor's theorem with an integral remainder (\cref{prop:Taylor}). Define $\bar{\alpha}_t(x)=\int_0^1(1-u)\dot{\mu}(x_t^\top\htt+u(x^\top\htt-x_t^\top\htt))du$, which accounts for higher-order terms based on the estimated parameter $\htt$ and feature vectors $x$ and $x_t$. The difference $\abr{\mu(\tx_t^\top\htt)-\mu(x_t^\top\htt)}$ then can be bounded as:
\begin{align*}
    \big|&\mu(\tx_t^\top\htt)-\mu(x_t^\top\htt)\big|=
    \abr{\dot{\mu}(x_t^\top\htt)\dotp{\tx_t-x_t}{\htt}+\int_{x_t^\top\htt}^{\tx_t^\top\htt}(\mu(\tx_t^\top\htt)-z)\ddot{\mu}(z) dz}\\
    &\leq\dot{\mu}(x_t^\top\htt)\abr{\dotp{\tx_t-x_t}{\htt}}+\dotp{\tx_t-x_t}{\htt}^2\int_0^1(1-u)\abr{\ddot{\mu}\rbr{x_t^\top\htt+u(\tx_t^\top\htt-x_t^\top\htt)}}du\\
    &\leq\dot{\mu}(x_t^\top\htt)\abr{\dotp{\tx_t-x_t}{\htt}}+M_\mu\dotp{\tx_t-x_t}{\htt}^2\underbrace{\int_0^1(1-u)\dot{\mu}\rbr{x_t^\top\htt+u(\tx_t^\top\htt-x_t^\top\htt)}du}_{=\bar{\alpha}_t(\tx_t)}\\
    &\leq\dot{\mu}(x_t^\top\htt)\abr{\dotp{\tx_t-x_t}{\htt}}+M_\mu\bar{\alpha}_t(\tx_t)\dotp{\tx_t-x_t}{\htt}^2\\
    &\leq\dot{\mu}(x_t^\top\htt)\gamma_t(\delta')\norm{x_t}_{\hat{H}_t^{-1}}+M_\mu\bar{\alpha}_t(\tx_t)\gamma_t(\delta')^2\norm{x_t}_{\hat{H}_t^{-1}}^2,\\
\end{align*}
where the second and the last inequalities hold from \cref{asm:Linkfunction} and the definition of $\Ecal_t(x_t)$ in \cref{def:events}. This bound captures both the linear and higher-order contributions to the regret from perturbations in the feature vectors.

\subparagraph{Bounding $A_t$} With $\ddot{E}_t := \cbr{\mu(\tx_t^\top\htt) \geq \mu(x_{t*}^\top\theta^*)}$ and $\tx_t^*$ defined above, we write:
\begin{align*}
    A_t&\ind{\hat{E}_t\cap\tilde{E}_t}
    =\rbr{\mu(x_{t*}^\top\theta^*)-\mu(\tx_t^\top\htt)}\ind{\hat{E}_t\cap\tilde{E}_t}\\
    &\leq\rbr{\mu(x_{t*}^\top\theta^*)-\inf_{\dbtilde{x}_t\in\Ecal_t(x_t)}\mu(\dbtilde{x}_t^\top\htt)}\ind{\hat{E}_t\cap\tilde{E}_t}\\
    &\leq\EE_t\sbr{\rbr{\mu(\tx_t^\top\htt)-\inf_{\dbtilde{x}_t\in\Ecal_t(x_t)}\mu(\dbtilde{x}_t^\top\htt)}\ind{\hat{E}_t\cap\tilde{E}_t}\bigg|\ddot{E}_t} \\  
    &=\EE_t\sbr{\rbr{\mu(\tx_t^\top\htt)-\mu(x_t^\top\htt)}+\rbr{\mu(x_t^\top\htt)-\inf_{\dbtilde{x}_t\in\Ecal_t(x_t)}\mu(\dbtilde{x}_t^\top\htt)}\bigg|\hat{E}_t,\tilde{E}_t,\ddot{E}_t}\PP(\hat{E}_t\cap\tilde{E}_t)\\
    &\leq2\EE_t\sbr{\rbr{\sup_{\dbtilde{x}_t\in\Ecal_t(x_t)}\abr{\mu(\dbtilde{x}_t^\top\htt)-\mu(x_t^\top\htt)}}\bigg|\hat{E}_t,\tilde{E}_t,\ddot{E}_t}\PP(\hat{E}_t\cap\tilde{E}_t) \\
    &\leq\frac{2}{p}\EE_t\sbr{\abr{\mu(\tx_t^{*\top}\htt)-\mu(x_t^\top\htt)}\ind{\hat{E}_t\cap\tilde{E}_t}}.
\end{align*}
We justify the second inequality under the specified event, and the final inequality follows from the following reasoning: we use the bound $C\leq\EE[Z\mid Z\geq C]$, and compensate for introducing the conditional expectation by incorporating the inverse of the probability of the conditioning event. Specifically, define $C:=(\mu(x_{t*}^\top\theta^*)-\inf_x\mu(x^\top\hat{\theta}_t)\cdot\ind(\hat{E}_t\cap\tilde{E}_t)$ and $Z:=(\mu(\tilde{x}_t^\top\htt)-\inf_x\mu(x^\top\hat{\theta}_t)\cdot\ind(\hat{E}_t\cap\tilde{E}_t)$. Then the second inequality holds. To compensate for conditioning on the favorable event, we use the following logic:
\begin{align*}
    \EE_t\cbr{\cdot\mid\hat{E}_t,\tilde{E}_t}&\geq\EE_t\cbr{\cdot\mid\hat{E}_t,\tilde{E}_t,\ddot{E}_t}\PP_t(\ddot{E}_t\mid\hat{E}_t,\tilde{E}_t)\\
    &\geq\EE_t\cbr{\cdot\mid\hat{E}_t,\tilde{E}_t,\ddot{E}_t}\cdot p.\tag{\cref{lem:anti_event}}
\end{align*}
The upper bound for this term is similar to the previous part ($B_t$), differing only by a constant and the inclusion of the expectation over the filtration. However, since our goal is to bound the cumulative sum over $T$ rounds rather than the expectation itself, directly handling the expectation complicates the application of the Elliptical Potential Lemma (EPL). To address this, we eliminate the expectation at the cost of introducing a concentration error, which we control using Azuma-Hoeffding's inequality.

\paragraph{Step 2-2 (Azuma-Hoeffding's inequality)} Unless otherwise specified, we now analyze the regret bound under the assumption that events $\hat{E}_T$ and $\tilde{E}_T$ hold.
\begin{align*}
    &\frac{p}{2}\sum_{t\in I_T}A_t\leq\sum_{t \in I_T}\EE_t\sbr{\abr{\mu(\tx_t^{*\top}\htt)-\mu(x_t^\top\htt)}}\\
    &\leq\gamma_T(\delta')\sum_{t\in I_T}\dot{\mu}(x_t^\top\htt)\norm{x_t}_{\hat{H}_t^{-1}}+\gamma_T(\delta')\underbrace{\sum_{t \in I_T}\rbr{\EE_t\sbr{\dot{\mu}(x_t^\top\htt)\norm{x_t}_{\hat{H}_t^{-1}}}-\dot{\mu}(x_t^\top\htt)\norm{x_t}_{\hat{H}_t^{-1}}}}_{R_1}\\
    &\quad+M_\mu\gamma_T(\delta')^2\sum_{t\in I_T}\bar{\alpha}_t(\tx_t)\norm{x_t}_{\hat{H}_t^{-1}}^2+M_\mu\gamma_T(\delta')^2\underbrace{\sum_{t \in I_T}\rbr{\EE_t\sbr{\bar{\alpha}_t(\tx_t)\norm{x_t}_{\hat{H}_t^{-1}}^2}-\bar{\alpha}_t(\tx_t)\norm{x_t}_{\hat{H}_t^{-1}}^2}}_{R_2}
\end{align*}
Note that $R_1$ and $R_2$ are constructed as martingales. Since the norm of each feature vector satisfies $\norm{x_t}\leq1$ , and given $\hat{H}_t^{-1}\preceq\hat{H}_0^{-1}=\Ib/\lambda$ and $\dot{\mu}(x_t^\top\htt)\leq\Lcal_\mu$, the following bounds hold:
\begin{align*}
    0\leq\dot{\mu}(x_t^\top\htt)\norm{x_t}_{\hat{H}_t^{-1}}\leq\Lcal_\mu\sqrt{x_t^\top\hat{H}_t^{-1}x_t}\leq\Lcal_\mu\sqrt{\frac{1}{\lambda}\norm{x_t}^2}&\leq\frac{\Lcal_\mu}{\sqrt{\lambda}}.
\end{align*}
This provides an upper bound for each instantaneous element of $R_1$ as $\Lcal_\mu/\sqrt{\lambda}$. Applying Azuma-Hoeffding's inequality (\cref{prop:Azuma's inequality}), with probability at least $1-\delta/4$, we obtain:
\begin{align*}
    R_1\leq\sqrt{\frac{2T\Lcal_\mu^2}{\lambda}\log\frac{8}{\delta}}.
\end{align*}
Due to the convexity of $\Theta_t$, the term $\dot{\mu}$ in $\bar{\alpha}_t(x_t)$ is bounded by $\Lcal_\mu$. Consequently, we obtain:
\begin{equation}\label{eqn:alpha}
    \bar{\alpha}_t(x):=\int_0^1(1-u)\dot{\mu}\rbr{x_t^\top\htt+u(x^\top\htt-x_t^\top\htt)}du\leq\int_0^1(1-u)\Lcal_\mu du=\Lcal_\mu/2
\end{equation}
Similarly, we can show that
\begin{align*}
    0\leq\bar{\alpha}_t(\tx_t)\norm{x_t}_{\hat{H}_t^{-1}}^2\leq\frac{\Lcal_\mu}{2}\rbr{x_t^\top\hat{H}_t^{-1}x_t}\leq\frac{\Lcal_\mu}{2\lambda}\norm{x_t}^2&\leq\frac{\Lcal_\mu}{2\lambda}.
\end{align*}
This provides an upper bound for each instantaneous element of $R_2$ as $\Lcal_\mu/(2\lambda)$. Applying \cref{prop:Azuma's inequality}, with probability at least $1-\delta/4$, we have:
\begin{align*}
    R_2\leq\sqrt{\frac{2T\Lcal_\mu^2}{4\lambda^2}\log\frac{8}{\delta}}.
\end{align*}
By applying a union bound, we conclude that with probability at least $1-\delta/2$, both $R_1$ and $R_2$ are bounded. Therefore, the regret term $\sum_{t\in I_T}A_t$ can be bounded by:
\begin{align*}
    \sum_{t\in I_T}A_t&\leq\frac{2}{p}\rbr{\gamma_T(\delta')\sum_{t\in I_T}\dot{\mu}(x_t^\top\htt)\norm{x_t}_{\hat{H}_t^{-1}}+M_\mu\gamma_T(\delta')^2\sum_{t\in I_T}\bar{\alpha}_t(\tx_t)\norm{x_t}_{\hat{H}_t^{-1}}^2}+\varepsilon,
\end{align*}
where $\varepsilon$ is a function of $T$, $\delta$, and $\lambda$, defined as:
\begin{align*}
    \varepsilon=\varepsilon(T,\delta,\lambda):=\frac{2}{p}\rbr{\gamma_t(\delta')\sqrt{\frac{2T\Lcal_\mu^2}{\lambda}\log\frac{8}{\delta}}+M_\mu\gamma_T(\delta')^2\sqrt{\frac{2T\Lcal_\mu^2}{4\lambda^2}\log\frac{8}{\delta}}}=\otil\rbr{d\sqrt{\frac{T}{\lambda^2}}}.
\end{align*}
In summary, combining this result with Lemma \ref{lem:anti_event}-\ref{lem:conc_event} and the bounds for $B_t$ and $C_t$, the total regret $\Reg(T)$ can be bounded with probability at least $1-\delta$ as:
\begin{align*}
    \Reg(T) \leq &\rbr{\frac{2}{p}+1}\underbrace{\sum_{t\in I_T}\rbr{\gamma_T(\delta')\dot{\mu}(x_t^\top\htt)\norm{x_t}_{\hat{H}_t^{-1}}+M_\mu\gamma_T(\delta')^2\bar{\alpha}_t(\tx_t)\norm{x_t}_{\hat{H}_t^{-1}}^2}}_{\Reg_\text{FP}} \\&+ \underbrace{\sum_{t \in I_T}\abr{\mu(x_t^\top\omega_t)-\mu(x_t^\top\htbt)}}_{\Reg_\text{EST}}+\varepsilon
\end{align*}
Here, $\Reg_\text{FP}$ captures the regret arising from perturbing the feature vectors, while $\Reg_\text{EST}$ accounts for the estimation error of $\htt$ compared to the true parameter $\theta^*$.

\paragraph{Step 3 (Bounding $\Reg_\text{FP}$)} The presence of $\hat{H}_t^{-1}$ in the weighted norm makes it challenging to directly apply the Elliptical Potential Lemma (EPL; \cref{lem:EPL}). To address this, we introduce $\bar{H}_t$, which allows us to leverage EPL by splitting $\dot{\mu}(x_t^\top\htt)$ into two components: a leading term based on $\btt$ (used in defining $\bar{H}_t$) and a transient term that accounts for deviations from $\htt$.
\begin{align*}
    &\Reg_\text{FP}=\gamma_T(\delta')\sum_{t\in I_T}\dot{\mu}(x_t^\top\htt)\norm{x_t}_{\hat{H}_t^{-1}}+M_\mu\gamma_T(\delta')^2\sum_{t\in I_T}\bar{\alpha}_t(\tx_t)\norm{x_t}_{\hat{H}_t^{-1}}^2\\
    &\leq\gamma_T(\delta')\sum_{t\in I_T}\rbr{\dot{\mu}(x_t^\top\btt)+\abr{\dot{\mu}(x_t^\top\btt)-\dot{\mu}(x_t^\top\htt)}}\norm{x_t}_{\hat{H}_t^{-1}}+M_\mu\gamma_T(\delta')^2\sum_{t\in I_T}\bar{\alpha}_t(\tx_t)\norm{x_t}_{\hat{H}_t^{-1}}^2\\
    &\leq\!\!\sum_{t\in I_T}\!\underbrace{\dot{\mu}(x_t^\top\btt)\gamma_T(\delta')\norm{x_t}_{\hat{H}_t^{-1}}}_{D_t^\text{FP}} \!+\!\!\sum_{t\in I_T}\!\underbrace{\abr{\dot{\mu}(x_t^\top\btt)\!-\!\dot{\mu}(x_t^\top\htt)}\gamma_T(\delta')\norm{x_t}_{\hat{H}_t^{-1}}}_{E_t^\text{FP}}\!+\!\!\sum_{t\in I_T}\!\underbrace{M_\mu\bar{\alpha}_t(\tx_t)\gamma_T(\delta')^2\norm{x_t}_{\hat{H}_t^{-1}}^2}_{F_t^\text{FP}}.
\end{align*}

\subparagraph{Bounding $\sum_t D_t^\text{FP}$}
We define $\bar{H}_t:=\lambda\Ib+\sum_{\tau=1}^{t-1}\dot{\mu}(x_\tau^\top\bar{\theta}_\tau)x_\tau x_\tau^\top$. Then for all $\tau\leq t$, as the equation $\dot{\mu}(x_\tau^\top\bar{\theta}_\tau)\leq\dot{\mu}(x_\tau^\top\htt)$ holds, we write:
\begin{align}\label{eqn:H_hat H_bar}
    \hat{H}_t=\lambda\Ib+\nabla^2\Lcal_t(\htt)=\lambda\Ib+\sum_{\tau=1}^{t-1}\dot{\mu}(x_\tau^\top\htt)x_\tau x_\tau^\top
    \succeq\lambda\Ib+\sum_{\tau=1}^{t-1}\dot{\mu}(x_\tau^\top\bar{\theta}_\tau)x_\tau x_\tau^\top\succeq\bar{H}_t.
\end{align}
Then, we can bound $\sum_t D_t^\text{FP}$ as following:
\begin{align*}
    \sum_{t\in I_T}D_t^\text{FP}&\leq\gamma_T(\delta')\sum_{t \in I_T}\dot{\mu}(x_t^\top\btt)\norm{x_t}_{\hat{H}_t^{-1}}\\
    &\leq \gamma_T(\delta')\sqrt{\sum_{t\in I_T}\dot{\mu}(x_t^\top\btt)}\sqrt{\sum_{t\in I_T}\dot{\mu}(x_t^\top\btt)\norm{x_t}_{\hat{H}_t^{-1}}^2} \tag{Cauchy-Schwartz inequality}\\
    &\leq \gamma_T(\delta')\sqrt{\sum_{t\in I_T}\dot{\mu}(x_t^\top\btt)}\sqrt{\sum_{t\in I_T}\dot{\mu}(x_t^\top\btt)\norm{x_t}_{\bar{H}_t^{-1}}^2} \tag{By \cref{eqn:H_hat H_bar}}\\
    &\leq \gamma_T(\delta')\sqrt{\sum_{t\in I_T}\dot{\mu}(x_t^\top\btt)}\sqrt{\sum_{t=1}^T\min\cbr{1,\norm{\sqrt{\dot{\mu}(x_t^\top\btt)}x_t}_{\bar{H}_t^{-1}}^2}} \tag{Definition of $I_T$}
\end{align*}
Applying \cref{lem:EPL}, the second square-root term is bounded by
\begin{align*}
    \sqrt{\sum_{t=1}^T\min\cbr{1,\norm{\sqrt{\dot{\mu}(x_t^\top\btt)}x_t}_{\bar{H}_t^{-1}}^2}}\leq\sqrt{2d\log\rbr{1+\frac{\Lcal_\mu T}{d\lambda}}}.
\end{align*}
To handle the first term in the square root, we decompose it as:
\begin{align*}
    \sqrt{\sum_{t\in I_T}\!\dot{\mu}(x_t^\top\btt)}\!\leq\!\sqrt{\sum_{t=1}^T\!\dot{\mu}(x_{t*}^\top\theta^*)\!+\!\sum_{t\in I_T}\!\cbr{\dot{\mu}(x_t^\top\btt)\!-\!\dot{\mu}(x_{t*}^\top\theta^*)}}\!=\!\sqrt{\kappa_*T\!+\!\sum_{t\in I_T}\!\cbr{\dot{\mu}(x_t^\top\btt)\!-\!\dot{\mu}(x_{t*}^\top\theta^*)}}.
\end{align*}
Note that $\btt:=\argmin_{\theta\in\cup_{\tau\in[t,T]}\Theta_\tau(\delta,\lambda)}\dot{\mu}(x_t^\top\theta)$. Let $\tau'$ be an arbitrary $\tau$ whose $\Theta_\tau(\delta,\lambda)$ contains $\btt$. Then, the latter term in the square root is then bounded as follows:
\begin{align*}
    &\sum_{t\in I_T}\cbr{\dot{\mu}(x_t^\top\btt)-\dot{\mu}(x_{t*}^\top\theta^*)} 
    =\sum_{t\in I_T}\cbr{\dot{\mu}(x_t^\top\btt)-\dot{\mu}(x_t^\top\theta^*)}+\sum_{t\in I_T}\cbr{\dot{\mu}(x_t^\top\theta^*)-\dot{\mu}(x_{t*}^\top\theta^*)}\\
    &\leq\! M_\mu\!\cbr{\sum_{t\in I_T}\abr{\mu(x_t^\top\btt)-\mu(x_t^\top\theta^*)}+\sum_{t\in I_T}\abr{\mu(x_t^\top\theta^*)-\mu(x_{t*}^\top\theta^*)}} \tag{\cref{lem:dot_mu mu}}\\
    &\leq \!M_\mu\!\cbr{\sum_{t\in I_T}\abr{\mu(x_t^\top\btt)-\mu(x_t^\top\hat{\theta}_{\tau'})}\!+\!\!\sum_{t\in I_T}\abr{\mu(x_t^\top\hat{\theta}_{\tau'})-\mu(x_t^\top\theta^*)}\!+\!\!\sum_{t\in I_T}\cbr{\mu(x_{t*}^\top\theta^*)-\mu(x_t^\top\theta^*)}\!}\\
    &\leq \!M_\mu\!\cbr{2\sum_{t\in I_T}\abr{\mu(x_t^\top\omega_t)-\mu(x_t^\top\htbt)}+\Reg(T)}\leq M_\mu \cbr{2\Reg_\text{EST}+\Reg(T)},
\end{align*}
where the second inequality holds from triangular inequality. For the last inequality, we leverage the definition of $\btt$ and the pair$(\tau(t),\omega_t)$, as shown below:
\begin{align*}
    \abr{\mu(x_t^\top\btt)-\mu(x_t^\top\hat{\theta}_{\tau'})}&\leq \max_{\theta\in\Theta_{\tau'}(\delta,\lambda)}\abr{\mu(x_t^\top\theta)-\mu(x_t^\top\hat{\theta}_{\tau'})}\\
    &\leq \max_{\tau\in[t,T],\theta\in\Theta_\tau(\delta,\lambda)}\abr{\mu(x_t^\top\theta)-\mu(x_t^\top\hat{\theta}_\tau)}=\abr{\mu(x_t^\top\omega_t)-\mu(x_t^\top\htbt)},
\end{align*}
and that under the event $\hat{E}_T$, as $\theta^*\in\cup_{\tau\in[t,T]}\Theta_\tau(\delta,\lambda)$, we can easily show that the second term $|\mu(x_t^\top\hat{\theta}_{\tau'})-\mu(x_t^\top\theta^*)|$ can be upper bounded by $|\mu(x_t^\top\omega_t)-\mu(x_t^\top\htbt)|$.
Note that after decomposing $\Reg(T)$, we are currently in the process of bounding its individual components. However, during this process, $\Reg(T)$ itself appears in the upper bound. This issue will be addressed in Step 5, where we will provide a strategy to effectively resolve this recursive dependency.

\subparagraph{Bounding $\sum_t E_t^\text{FP}$} We define the Gram matrix $\bar{V}_t:=\lambda\Ib/\kappa+\sum_{\tau=1}^{t-1}x_\tau x_\tau^\top$. The introduction of this matrix is to apply EPL as in the previous section. As the weight $\dot{\mu}(x_\tau^\top\htt)$ on the weighted Gram matrix $\hat{H}_t$ is greater than or equal to $\kappa:=\min_{x\in\Xcal_{[T]},\theta\in\Theta}\dot{\mu}(x^\top\theta)$, we have:
\begin{align}\label{eqn:H_hat V}
    \hat{H}_t=\lambda\Ib+\sum_{\tau=1}^{t-1}\dot{\mu}(x_\tau^\top\htt)x_\tau x_\tau^\top\succeq\kappa\rbr{\frac{\lambda}{\kappa}\Ib+\sum_{\tau=1}^{t-1}x_\tau x_\tau^\top}\succeq\kappa\bar{V}_t.
\end{align}
Then, we can bound $\sum_t E_t^\text{FP}$ as following:
\begin{align*}
    \sum_{t\in I_T} E_t^\text{FP}&\leq\gamma_T(\delta')\sum_{t\in I_T}\abr{\dot{\mu}(x_t^\top\btt)-\dot{\mu}(x_t^\top\htt)}\norm{x_t}_{\hat{H}_t^{-1}}\\
    &\leq M_\mu\gamma_T(\delta')\sum_{t\in I_T}\abr{\mu(x_t^\top\btt)-\mu(x_t^\top\htt)}\norm{x_t}_{\hat{H}_t^{-1}} \tag{\cref{lem:dot_mu mu}}\\
    &\leq M_\mu\gamma_T(\delta')\sqrt{\frac{1}{\kappa}}\sum_{t\in I_T}\abr{\mu(x_t^\top\omega_t)-\mu(x_t^\top\htbt)}\norm{x_t}_{\bar{V}_t^{-1}}\\
    &\leq \frac{2}{\kappa}M_\mu\Lcal_\mu\gamma_T(\delta')\beta_T(\delta')\sum_{t\in I_T}\norm{x_t}_{\bar{V}_t^{-1}}^2,\tag{\cref{lem:mu diff}}
\end{align*}
where the third inequality holds from the definition of the pair $(\tau(t),\omega_t)$, and by \cref{eqn:H_hat V}
Here, \cref{lem:mu diff} plays a crucial role by introducing the weighted norm $\norm{x_t}_{\bar{V}_t^{-1}}$, which enables the application of \cref{lem:EPL}. Utilizing this lemma, we further proceed to bound as:
\begin{align*}
    \sum_{t\in I_T} E_t^\text{FP}&\leq \frac{2}{\kappa}M_\mu\Lcal_\mu\gamma_T(\delta')\beta_T(\delta')\sum_{t=1}^T\min\cbr{1,\norm{x_t}_{\bar{V}_t^{-1}}^2}\tag{Definition of $I_T$}\\
    &\leq \frac{4}{\kappa}dM_\mu\Lcal_\mu\gamma_T(\delta')^2\log\rbr{1+\frac{\kappa T}{d\lambda}} \tag{\cref{lem:EPL}},
\end{align*}
where the final inequality use the fact that $\gamma_T(\delta')=c_{\delta'}\cdot\beta_T(\delta')\geq\beta_T(\delta')$, as $c_{\delta'}\geq1$.

\subparagraph{Bounding $\sum_t F_t^\text{FP}$} The process closely resembles that of bounding $\sum_t E_t^\text{FP}$. we have:
\begin{align*}
    \sum_{t\in I_T} F_t^\text{FP}&\leq M_\mu\gamma_T(\delta')^2\sum_{t\in I_T}\bar{\alpha}_t(\tx_t)\norm{x_t}_{\hat{H}_t^{-1}}^2\\
    &\leq \frac{\Lcal_\mu}{2\kappa}M_\mu\gamma_T(\delta')^2\sum_{t\in I_T}\norm{x_t}_{\bar{V}_t^{-1}}^2 \tag{By \cref{eqn:alpha} and \cref{eqn:H_hat V}}\\
    &\leq \frac{\Lcal_\mu}{2\kappa}M_\mu\gamma_T(\delta')^2\sum_{t=1}^T\min\cbr{1,\norm{x_t}_{\bar{V}_t^{-1}}^2} \tag{Definition of $I_T$}\\
    &\leq \frac{d}{\kappa} M_\mu\Lcal_\mu\gamma_T(\delta')^2\log\rbr{1+\frac{\kappa T}{d\lambda}}.\tag{\cref{lem:EPL}}
\end{align*}

Combining all three terms, we derive an upper bound for $\Reg_{\text{FP}}\leq\sum_t \cbr{D_t^\text{FP}+E_t^\text{FP}+F_t^\text{FP}}$ as follows:
\begin{align}
    \Reg_{\text{FP}}&\leq \gamma_T(\delta')\sqrt{2d\log\rbr{1+\frac{\Lcal_\mu T}{d\lambda}}}\sqrt{\kappa_*T+M_\mu \cbr{2\Reg_\text{EST}+\Reg(T)}}\notag\\
    &\quad+ \frac{5d}{\kappa} M_\mu\Lcal_\mu\gamma_T(\delta')^2\log\rbr{1+\frac{\kappa T}{d\lambda}}.\label{eqn:reg_FP}
\end{align}

\paragraph{Step 4 (Bounding $\Reg_\text{EST}$)} Next, we proceed to bound $\Reg_{\text{EST}}$. The overall process closely mirrors that of $\Reg_{\text{FP}}$. To begin, we decompose each instantaneous regret term. Let $\bar{\alpha}_t(\theta_1,\theta_2)=\int_0^1(1-u)\dot{\mu}(x_t^\top\theta_1+u(x_t^\top\theta_2-x_t^\top\theta_1))du$, which is derived from Taylor's theorem when estimating the regret for the selected arm $x_t$ using two parameter vectors, $\theta_1$ and $\theta_2$. Using \cref{prop:Taylor}, we expand the instantaneous regret term as:
\begin{align*}
    &\big|\mu(x_t^\top\omega_t)-\mu(x_t^\top\htbt)\big|=
    \abr{\dot{\mu}(x_t^\top\htbt)\dotp{x_t}{\omega_t-\htbt}+\int_{x_t^\top\htbt}^{x_t^\top\omega_t}(\mu(x_t^\top\omega_t)-z)\ddot{\mu}(z) dz}\\
    &\leq\dot{\mu}(x_t^\top\htbt)\abr{\dotp{x_t}{\omega_t-\htbt}}+\dotp{x_t}{\omega_t-\htbt}^2\int_0^1(1-u)\abr{\ddot{\mu}\rbr{x_t^\top\htbt+u(x_t^\top\omega_t-x_t^\top\htbt)}}du\\
    &\leq\dot{\mu}(x_t^\top\htt)\abr{\dotp{x_t}{\omega_t-\htbt}}+M_\mu\dotp{x_t}{\omega_t-\htbt}^2\underbrace{\int_0^1(1-u)\dot{\mu}\rbr{x_t^\top\htbt+u(x_t^\top\omega_t-x_t^\top\htbt)}du}_{=\bar{\alpha}_t(\htbt,\omega_t)}\\
    &\leq\rbr{\dot{\mu}(x_t^\top\btt)+\abr{\dot{\mu}(x_t^\top\btt)-\dot{\mu}(x_t^\top\htbt)}}\abr{\dotp{x_t}{\omega_t-\htbt}}+M_\mu\bar{\alpha}_t(\htbt,\omega_t)\dotp{x_t}{\omega_t-\htbt}^2\\
    &\leq\rbr{\dot{\mu}(x_t^\top\btt)+\abr{\dot{\mu}(x_t^\top\btt)-\dot{\mu}(x_t^\top\htbt)}}\abr{\dotp{x_t}{\omega_t-\htbt}}+\underbrace{M_\mu\bar{\alpha}_t(\htbt,\omega_t)\beta_{\tau(t)}(\delta')^2\norm{x_t}_{\hat{H}_{\tau(t)}^{-1}}^2}_{F_t^\text{EST}},
\end{align*}
where the second inequality follows from \cref{asm:Linkfunction}, and the last inequality results from the Cauchy-Schwartz inequality. The key distinction between bounding $\Reg_\text{FP}$ and $\Reg_\text{EST}$ lies is the use of $\beta(\delta')$ instead of $\gamma(\delta')$. The first term then can be upper bounded by the following terms, using triangular inequality:
\begin{align*}
    \Big(\dot{\mu}(x_t^\top\btt)+\Big|\dot{\mu}(x_t^\top&\btt)-\dot{\mu}(x_t^\top\htbt)\Big|\Big)\norm{x_t}_{\hat{H}_{\tau(t)}^{-1}}\norm{\omega_t-\htbt}_{\hat{H}_{\tau(t)}}\\
    &\leq\underbrace{\dot{\mu}(x_t^\top\btt)\beta_{\tau(t)}(\delta')\norm{x_t}_{\hat{H}_{\tau(t)}^{-1}}}_{D_t^\text{EST}}+\underbrace{\abr{\dot{\mu}(x_t^\top\btt)-\dot{\mu}(x_t^\top\htt)}\beta_{\tau(t)}(\delta')\norm{x_t}_{\hat{H}_{\tau(t)}^{-1}}}_{E_t^\text{EST}}.
\end{align*}

\subparagraph{Bounding $\sum_t D_t^\text{EST}$} By the equation\cref{eqn:H_hat H_bar}, instead of $t$-dependent $\hat{H}_t$, we use $\tau$-only dependent $\bar{H}_t$ to upper bound each term. We have:
\begin{align*}
    \sum_{t\in I_T}D_t^\text{EST} &\leq\beta_T(\delta')\sum_{t\in I_T}\dot{\mu}(x_t^\top\btt)\norm{x_t}_{\hat{H}_t^{-1}} \tag{$t\leq\tau(t)\leq T$}\\
    &\leq \beta_T(\delta')\sqrt{2d\log\rbr{1+\frac{\Lcal_\mu T}{d\lambda}}}\sqrt{\sum_{t=1}^T\dot{\mu}(x_{t*}^\top\theta^*)+\sum_{t\in I_T}\cbr{\dot{\mu}(x_t^\top\btt)-\dot{\mu}(x_{t*}^\top\theta^*)}} \\
    &\leq\beta_T(\delta')\sqrt{2d\log\rbr{1+\frac{\Lcal_\mu T}{d\lambda}}}\sqrt{\kappa_*T+M_\mu \cbr{2\Reg_\text{EST}+\Reg(T)}}.
\end{align*}
Since the derivation proceeds identically to the case of $\sum_t D_t^{\text{FP}}$, with the only difference being the use of $\beta_T$ in place of $\gamma_T$, we omit the detailed derivation here for brevity.

\subparagraph{Bounding $\sum_t E_t^\text{EST}$} Following the bounding process of $\sum_t E_t^\text{FP}$, and using the equation \cref{eqn:H_hat V},
\begin{align*}
    \sum_{t\in I_T} E_t^\text{EST}&\leq\beta_T(\delta')\sum_{t\in I_T}\abr{\dot{\mu}(x_t^\top\btt)-\dot{\mu}(x_t^\top\htt)}\norm{x_t}_{\hat{H}_t^{-1}} \tag{$t\leq\tau(t)\leq T$}\\
    &\leq \frac{4d}{\kappa}M_\mu\Lcal_\mu\beta_T(\delta')^2\log\rbr{1+\frac{\kappa T}{d\lambda}}.
\end{align*}

\subparagraph{Bounding $\sum_t F_t^\text{EST}$} For $\bar{\alpha}_t(\htbt,\omega_t):=\int_0^1(1-u)\dot{\mu}(x_t^\top\htbt+u(x_t^\top\omega_t-x_t^\top\htbt))du$, as shown in \cref{eqn:alpha}, it follows that $\bar{\alpha}_t(\htbt,\omega_t)\leq\Lcal_\mu/2$. Following the bounding process of $\sum_t F_t^\text{FP}$,
\begin{align*}
    \sum_{t\in I_T} F_t^\text{EST}&\leq M_\mu\beta_T(\delta')^2\sum_{t\in I_T}\bar{\alpha}_t(\tx_t)\norm{x_t}_{\hat{H}_t^{-1}}^2 \tag{$t\leq\tau(t)\leq T$}\\
    &\leq \frac{d}{\kappa} M_\mu\Lcal_\mu\beta_T(\delta')^2\log\rbr{1+\frac{\kappa T}{d\lambda}}.
\end{align*}

In summary, combining three regret components, we bound $\Reg_{\text{EST}}\leq\sum_t \cbr{D_t^\text{EST}+E_t^\text{EST}+F_t^\text{EST}}$ as follows:
\begin{align}
    \Reg_{\text{FP}}&\leq \beta_T(\delta')\sqrt{2d\log\rbr{1+\frac{\Lcal_\mu T}{d\lambda}}}\sqrt{\kappa_*T+M_\mu \cbr{2\Reg_\text{EST}+\Reg(T)}}\notag\\
    &\quad+ \frac{5d}{\kappa} M_\mu\Lcal_\mu\beta_T(\delta')^2\log\rbr{1+\frac{\kappa T}{d\lambda}}\label{eqn:reg_EST}.
\end{align}

\paragraph{Step 5 (Solving equation)}
The equation \cref{eqn:reg_EST} is indeed $c_{\delta'}>1$ times the equation \cref{eqn:reg_FP}. Therefore, both $\Reg_\text{FP}$ and $\Reg_\text{EST}$ can be bounded using the regret bound of $\Reg_\text{FP}$. Let $\Reg_{\max}=\max\cbr{\Reg_\text{FP},\Reg_\text{EST}}$, and we can bound $\Reg_{\max}$ with the equation \cref{eqn:reg_FP}. Then we have,
\begin{align*}
    \Reg(T) &\leq \rbr{\frac{2}{p}+1}\Reg_\text{FP}+\Reg_\text{EST}+\varepsilon\leq\rbr{\frac{2}{p}+2}\Reg_{\max}+\varepsilon,
\end{align*}
and accordingly, we get the following inequality:
\begin{align*}
    \Reg_{\max}&\leq \gamma_T(\delta')\sqrt{2d\log\rbr{1+\frac{\Lcal_\mu T}{d\lambda}}}\sqrt{\kappa_*T+M_\mu \cbr{\rbr{\frac{2}{p}+4}\Reg_{\max}+\varepsilon}}\\
    &\quad+ \frac{5d}{\kappa} M_\mu\Lcal_\mu\gamma_T(\delta')^2\log\rbr{1+\frac{\kappa T}{d\lambda}}
\end{align*}
Now, The bound takes the form of $\Reg_{\max}\leq A\sqrt{B+C\Reg_{\max}}+D$, where the following quantities are defined:
\begin{align*}
    A&:=\gamma_T(\delta')\sqrt{2d\log\rbr{1+\frac{\Lcal_\mu T}{d\lambda}}}=\otil\rbr{d}\\
    B&:=\kappa_*T+M_\mu\varepsilon=\otil\rbr{\kappa_*T+d\sqrt{\frac{T}{\lambda^2}}}\\
    C&:=M_\mu\rbr{\frac{2}{p}+4}=\Ocal(1)\\
    D&:=\frac{5d}{\kappa} M_\mu\Lcal_\mu\gamma_T(\delta')^2\log\rbr{1+\frac{\kappa T}{d\lambda}}=\otil\rbr{\frac{d^2}{\kappa}},
\end{align*}
focusing on the terms involving $d$,$T$, and $\kappa$. With the choice of $\lambda=\Ocal(d)$ and applying \cref{lem:poly solve}, the upper bound for $\Reg_{\max}$ simplifies to:
\begin{align*}
    \Reg_{\max}&=\otil\rbr{d\sqrt{\kappa_*T}+d^2+\frac{d^2}{\kappa}}=\otil\rbr{d\sqrt{\kappa_*T}+\frac{d^2}{\kappa}}.
\end{align*}
Now, combining all terms, the cumulative regret $R(T)$ can be bounded as:
\begin{align*}
    R(T) &\leq \Reg(T) + \frac{4d R^*}{\log 2}\cbr{\log\rbr{1+\frac{\Lcal_\mu}{\lambda\log2}}+\log\rbr{1+\frac{\kappa}{\lambda\log2}}}\\
    &\lesssim\rbr{\frac{2}{p}+2}\Reg_{\max}+\varepsilon\\
    &=\otil\rbr{d\sqrt{\kappa_*T}+\frac{d^2}{\kappa}+d\sqrt{\frac{T}{\lambda^2}}}=\otil\rbr{d\sqrt{\kappa_*T}+\frac{d^2}{\kappa}}.\tag{Choose $\lambda=\Ocal(d)$}
\end{align*}
Our derived regret bound, $\otil\rbr{d\sqrt{\kappa_*T}+d^2/\kappa}$, aligns with the state-of-the-art regret guarantee~\citep{Abeille2020InstanceWiseMA, Faury2020Improveds, Lee2024Improvedlog, lee2025unified}.

\newpage
\section{Proof supplement and Lemmas}\label{app:supporting lemma}
In this section, we provide key propositions, lemmas, and inequality bounds that are essential for the main proof. These results serve as the mathematical foundation for the regret analysis and other theoretical guarantees established in this work.

\subsection{Supporting Lemmas for main proof}
\begin{proposition}[Lemma D.1. of \citet{lee2025unified}]\label{prop:tildeH to nabla} Let $\mu$ be increasing and self-concordant with $M_\mu$. Let $\Zcal\subseteq\Bcal(S):=\cbr{z\in\RR~|~\abr{z}\leq S}$ in $\RR$. Then for any $z_1,z_2\in\Zcal$, the following holds:
\begin{align*}
    \int_0^1(1-u)\dot{\mu}(z_1+u(z_2-z_1))du\geq\frac{\dot{\mu}(z_1)}{2+2SM_\mu}
\end{align*}
\end{proposition}
This proposition establishes a lower bound for the integral involving a self-concordant, particularly useful for controlling weighted norms and derivatives in regret analysis. The following proposition provides the well-known Taylor's expansion expression with integral remainder. We highlight specific cases that are frequently used throughout the main proof.
\begin{proposition}[Taylor's Theorem with Integral Remainder Form]\label{prop:Taylor}
    Let $n\geq0$ be an integer and let the function $f:\RR\rightarrow\RR$ be $(n+1)$ times differentiable at the point $x_0\in\RR$. Let $f^{(n)}$ to denote its $n$-th derivatives, then $f(x)$ can be expressed as:
    \begin{align*}
        f(x)=\sum_{i=0}^n\frac{f^{(i)}(x_0)}{i!}(x-x_0)^i+\frac{1}{n!}\int_{x_0}^x f^{(n+1)}(t)(x-t)^ndt\label{eqn:taylor_0}
    \end{align*}
Especially for $n=0$, by letting $t=x_0+u(x-x_0)$, $f(x)$ can be expressed as:
\begin{align}
    f(x)&=f(x_0)+\int_{x_0}^xf'(t)dt=f(x_0)+(x-x_0)\int_0^1 f'\rbr{x_0+u(x-x_0)}du
\end{align}
Especially for $n=1$, by letting $t=x_0+u(x-x_0)$, $f(x)$ can be expressed as:
\begin{align}
    f(x)&=f(x_0)+f'(x_0)(x-x_0)+\int_{x_0}^xf''(t)(x-t)dt \nonumber\\
    &=f(x_0)+f'(x_0)(x-x_0)+\int_0^1 f''\rbr{x_0+u(x-x_0)}((1-u)(x-x_0))\cdot(x-x_0)du \nonumber\\
    &=f(x_0)+f'(x_0)(x-x_0)+(x-x_0)^2\int_0^1 f''\rbr{x_0+u(x-x_0)}(1-u)du\label{eqn:taylor_1}
\end{align}
Similarly, with multivariate function $f:\RR^d\rightarrow\RR$ and $n=1$, we have that:
\begin{align*}
    f(x)&=f(x_0)+\sum_{i=1}^d\frac{\partial}{\partial x_i} f(x_0)(x-x_0)_i+\sum_{i=1}^d(x-x_0)_i^2\int_0^1(1-u)\frac{\partial^2}{\partial x_i^2}f(x_0+u(x-x_0))du\\
    &\quad+\sum_{i\not=j}2(x-x_0)_i(x-x_0)_j\int_0^1(1-u)\frac{\partial}{\partial x_i}\frac{\partial}{\partial x_j}f(x_0+u(x-x_0))du\\
    &=f(x_0)\!+\!\nabla f(x_0)^\top(x-x_0)\!+\!(x-x_0)^\top\rbr{\int_0^1(1-u)\nabla^2f(x_0+u(x-x_0))du}(x-x_0).
\end{align*}
\end{proposition}
The following lemmas provide critical tools for analyzing regret bounds in the context of bandit algorithms. Specifically, they focus on bounding terms that involve weighted norms of feature vectors under Gram matrices. These results are pivotal in deriving efficient bounds on cumulative regret and exploration.
\begin{lemma}[Elliptical Potential Lemma]\label{lem:EPL}
    let $\cbr{x_t}_{t=1}^T$ be a sequence in $\RR^d$ satisfying $\norm{x_t}\leq R$ for all $t\leq T$. For a gram matrix $V_t:=\lambda\Ib+\sum_{\tau=1}^{t-1}x_\tau x_\tau^\top$, we have that
    \begin{equation*}
        \sum_{t=1}^T\min\cbr{1,\norm{x_t}_{V_t^{-1}}^2}\leq 2d\log\rbr{1+\frac{R^2T}{d\lambda}}.
    \end{equation*}
\end{lemma}

\begin{lemma}[Elliptical Potential Count Lemma]\label{lem:EPCL}
    For $R,c>0$, let $\cbr{x_t}_{t=1}^T$ be a sequence in $\RR^d$ satisfying $\norm{x_t}\leq R$ for all $t\leq T$. For a gram matrix $V_t:=\lambda\Ib+\sum_{\tau=1}^{t-1}x_\tau x_\tau^\top$, the length of the sequence $\Ncal_T:=\cbr{t\in[T]~|~\norm{x_t}_{V_t^{-1}}>c}$ is bounded as:
    \begin{equation*}
        \abr{\Ncal_T}\leq\frac{2d}{\log(1+c^2)}\log\rbr{1+\frac{R^2}{\lambda\log(1+c^2)}}
    \end{equation*}
\end{lemma}

The following connects the smoothness of $\dot{\mu}$ to the difference in $\mu$.
\begin{lemma}\label{lem:dot_mu mu}
    For $x,y\in\RR$, $\abr{\dot{\mu}(x)-\dot{\mu}(y)}\leq M_\mu\abr{\mu(x)-\mu(y)}$.
\end{lemma}
\begin{proof}
    \begin{align*}
        \abr{\dot{\mu}(x)-\dot{\mu}(y)}&=\abr{(x-y)\int_0^1\ddot{\mu}(y+u(x-y))du} \tag{By \cref{eqn:taylor_0} with $f=\dot{\mu}$}\\
        &\leq\abr{x-y}\int_0^1\abr{\ddot{\mu}(y+u(x-y))}du\\
        &\leq M_\mu\abr{x-y}\int_0^1\dot{\mu}(y+u(x-y))du \tag{\cref{asm:Linkfunction}}\\
        &=M_\mu\abr{(x-y)\int_0^1\dot{\mu}(y+u(x-y))du}\\
        &=M_\mu\abr{\mu(x)-\mu(y)} \tag{By \cref{eqn:taylor_0} with $f=\mu$}
    \end{align*}
    where the second equality holds since the integral term is positive as $\dot{\mu}\geq0$.
\end{proof}

The following lemma bounds the difference in $\mu$ values, using linearization as in the previous works.
\begin{lemma}\label{lem:mu diff}
    For any $t\geq1$ and $\theta_1,\theta_2\in\Theta_{\tau(t)}$, we have the following:
    \begin{equation*}
        \abr{\mu(x_t^\top\theta_1)-\mu(x_t^\top\theta_2)}\leq2\gamma_T(\delta)\Lcal_\mu\sqrt{\frac{1}{\kappa}}\norm{x_t}_{V_t^{-1}}
    \end{equation*}
\end{lemma}
\begin{proof}
    \begin{align*}
        |\mu(x_t^\top\theta_1)&-\mu(x_t^\top\theta_2)|=\abr{\dotp{x_t}{\theta_1-\theta_2}\int_0^1\dot{\mu}(x_t^\top\theta_2+u(x_t^\top\theta_1-x_t^\top\theta_2))du}\tag{By \cref{eqn:taylor_0}}\\
        &\leq\cbr{\norm{x_t}_{\hat{H}_{\tau(t)}^{-1}}\cdot\norm{\theta_1-\theta_2}_{\hat{H}_{\tau(t)}}}\int_0^1\Lcal_\mu du\tag{Cauchy-Schwartz, $\Lcal_\mu$-Lipschitzness}\\
        &\leq\Lcal_\mu\norm{x_t}_{\hat{H}_{\tau(t)}^{-1}}\cbr{\norm{\theta_1-\htbt}_{\hat{H}_{\tau(t)}}+\norm{\theta_2-\htbt}_{\hat{H}_{\tau(t)}}}\tag{Triangle inequality}\\
        &\leq2\beta_{\tau(t)}(\delta')\Lcal_\mu\norm{x_t}_{\hat{H}_{\tau(t)}^{-1}}\\
        &\leq2\beta_{\tau(t)}(\delta')\Lcal_\mu\sqrt{\frac{1}{\kappa}}\norm{x_t}_{V_{\tau(t)}^{-1}}\tag{By \cref{eqn:H_hat V}}\\
        &\leq2\beta_T(\delta')\Lcal_\mu\sqrt{\frac{1}{\kappa}}\norm{x_t}_{V_t^{-1}}\tag{$t\leq\tau(t)\leq T$}
    \end{align*}
\end{proof}

\begin{lemma}\label{lem:poly solve}
    Let $A,B,C,D,X\in\RR^+$. The following implication holds:
    \begin{equation*}
        X\leq A\sqrt{B+CX}+D\quad\Longrightarrow\quad X\leq 2(A\sqrt{B}+A^2C+D)
    \end{equation*}
\end{lemma}
\begin{proof}
    let $f:x\rightarrow x^2-px-q$ for $p,q>0$. Then the roots for $f(x)=0$ are:
    \begin{equation*}
        x_1,x_2=\frac{p\pm\sqrt{p^2+4q}}{2}.
    \end{equation*}
    Now, as $f$ is a convex function, $x^2\leq px+q$ implies:
    \begin{equation}
        x\leq\max\cbr{x_1,x_2}\leq\frac{p+\sqrt{p^2+4q}}{2}\leq\frac{p+(p+2\sqrt{q})}{2}=p+\sqrt{q}.\tag{triangle inequality}
    \end{equation}
    And accordingly, we have:
    \begin{align}
        x\leq p\sqrt{x}+q&\quad\Longrightarrow\quad \sqrt{x}\leq p+\sqrt{q}\notag\\
        &\quad\Longrightarrow\quad x\leq (p+\sqrt{q})^2\leq2p^2+2q\label{eqn:polynomial inequ}
    \end{align}
    where the inequality holds from $(x+y)^2\leq2(x^2+y^2)$. Then according to the equation \cref{eqn:polynomial inequ},
    \begin{align*}
        X\leq A\sqrt{B+CX}+D&\quad\Longrightarrow\quad X\leq A\sqrt{C}\sqrt{X}+A\sqrt{B}+D\tag{triangle inequality}\\
        &\quad\Longrightarrow\quad X\leq 2(A\sqrt{C})^2+2(A\sqrt{B}+D)=2(A\sqrt{B}+A^2C+D)
    \end{align*}
\end{proof}

\subsection{Auxiliary bounding inequalities}
We introduce key probabilistic inequalities and bounds frequently used in the analysis of randomized algorithms. These results provide tools to bound the probabilities of deviations and concentration of random variables, which are essential for deriving high-probability guarantees in the main analysis.
\begin{proposition}[Azuma's inequality] \label{prop:Azuma's inequality}
    If a super-martingale $(X_t)_{t\geq0}$ corresponding to a filtration $\Hcal_{t-1}$ satisfies $|X_t-X_{t-1}|<c_t$ for some constant $c_t$ for all $t=1,\ldots,T$ then for any $\alpha>0$,
    \begin{equation*}
        \PP(X_T-X_0\geq\alpha)\leq2\exp\rbr{-\frac{\alpha^2}{2\sum_{t=1}^T c_t^2}}.
    \end{equation*}
\end{proposition}
\begin{proposition}[Chernoff bound] \label{prop:Chernoff}
    For a random variable $X$ and its moment-generating function $M(t)=\EE\sbr{e^{tX}}$,
    \begin{equation*}
        \PP(X\geq \alpha)\leq\inf_{t>0}M(t)e^{-t\alpha}.
    \end{equation*}
    Accordingly, for a random variable following a standard normal distribution (i.e. $X\sim \Ncal(0,1)$),
    \begin{equation*}
            \PP(X\geq\alpha)\leq\inf_{t>0}\exp(\frac{t^2}{2}-t\alpha)=e^{-\frac{\alpha^2}{2}}.
    \end{equation*}
\end{proposition}
\begin{lemma}\label{lem:normal bound}
    Let $z$ be a random variable sampled from the standard Normal distribution. Then for all $\delta\in(0,1)$,
    \begin{equation*}
        \PP(|z|\leq\sqrt{2\log(2/\delta)})\geq 1-\delta.
    \end{equation*}
\end{lemma}
\begin{proof}
    By proposition \pref{prop:Chernoff}, $\PP(|z|>\alpha)=2\PP(z>\alpha)\leq2e^{-\frac{\alpha^2}{2}}$. Set the right-hand side as $\delta$, and get $\alpha=\sqrt{2\log(2/\delta)}$.
\end{proof}

\section{Limitations}\label{sec:limitation}
Our analysis focuses on structured settings such as generalized linear bandits (GLBs), where feature perturbation achieves both strong empirical performance and provable regret guarantees. 
While the same principle shows promise in more flexible or non-linear models, theoretical guarantees in these broader settings remain open. 
The current formulation, though practically effective, is heuristic outside the GLM framework and lacks formal justification under complex function classes. 
Extending the theory to overparameterized or general Lipschitz models represents an important direction for future work, where feature-level stochasticity may offer a stable alternative to parameter perturbation.
\newpage
\section{Experimental settings and additional results}\label{app:additional exp}
\subsection{Experimental details}
The GLB experiments are entirely synthetic and computationally lightweight. All GLB runs were performed on a standard CPU server equipped with an Intel Xeon Silver 4210R processor (40 threads), with each run completing within a few minutes. The neural bandit experiments were conducted using a single NVIDIA RTX 3090 GPU. Due to the moderate model size and limited data horizon, neural runs for each algorithm completed in under one hour. Overall, the compute requirements were modest, and no large-scale pretraining or extensive hyperparameter tuning was necessary for any of the experiments.

\subsubsection{Generalized linear bandit settings}
We consider a time-varying set of $K = 100$ arms per round over a horizon of $T = 20{,}000$ (linear) or $T = 10{,}000$ (logistic). Context vectors and the true parameter vector $\theta^*$ are sampled from a standard multivariate Gaussian distribution and normalized to satisfy the boundedness assumption. In the logistic case, we use the sigmoid link function $\mu(x) = 1 / (1 + e^{-x})$ and constrain $\|\theta^*\| \leq 4$ so that the logits lie in $[-4, 4]$. For the linear bandit, the reward noise is sampled from $\Ncal(0, 1)$. We vary the feature dimension $d$ over $\{10, 20, 40\}$. The confidence level is set as $\delta = 1/T$, and the regularization parameter is fixed at $\lambda = 10^{-4}$.

\paragraph{Baselines and tuning.}
For linear bandits, we compare against $\varepsilon$-greedy~\cite{langford2007epsilon}, LinUCB~\cite{abbasi2011improved}, LinTS~\cite{agrawal2013thompson}, LinPHE~\cite{kveton2020perturbed}, and RandLinUCB~\cite{vaswani20old}. For logistic bandits, we include $\varepsilon$-greedy, OFUL-GLB-e~\cite{lee2025unified}, LogTS~\cite{kveton2020randomized}, LogPHE~\cite{kveton2020randomized}, and RandUCBLog. All hyperparameters, such as inflation factors and learning rates, are tuned according to the original papers. We fix $\Lcal_\mu = 0.25$ and use a minimum link derivative of $0.25$ for numerical stability in the logistic setting. For TS-based algorithms and our method, we set the inflation parameter as $c_t = 1$. For $\varepsilon$-greedy, we use an annealing schedule $\varepsilon_t = \varepsilon \sqrt{T/t}$ with $\varepsilon = 0.05$.

\subsubsection{Neural bandit settings}

\paragraph{Objective.}
Following \citet{neuralts}, we evaluate neural contextual bandits using classification tasks on UCI benchmark datasets~\cite{Dua:2017}: \texttt{shuttle}, \texttt{isolet}, and \texttt{mushroom}. These tasks are transformed into multi-armed bandit problems via a disjoint model construction (cf. \citet{li2010contextual}). For a $k$-class classification problem with input dimension $d$, we construct a $kd$-dimensional feature representation by placing the input vector $x$ into the corresponding class slot: $x_1 = (x;\zero;\cdots;\zero)$, $x_2 = (\zero;x;\cdots;\zero)$, and so on. Each $x_i$ is treated as the feature vector for the $i$-th arm. A neural model $f$ predicts the reward for each $x_i$, and the agent selects the arm with the highest predicted reward. A reward of 1 is given if the selected arm corresponds to the correct label, and 0 otherwise. Regret is measured as the cumulative number of classification errors. All experiments are repeated five times with shuffled data. The time horizon is set to $T = 10{,}000$ for all datasets.

\paragraph{Neural networks.}
To study the effect of model capacity, we evaluate two neural network architectures: a shallow and a deep model. The shallow network consists of a single hidden layer with a fully connected layer of size $d \times 100$, followed by a ReLU activation, a final fully connected layer of size $100 \times 1$, and a softmax output layer, following the design used in \citet{neuralts}. The deep network, in contrast, uses two hidden layers with a fully connected layer of size $d \times 50$, a ReLU activation, another fully connected layer of size $50 \times 50$ followed by ReLU, and a final output layer of size $50 \times 1$ with softmax. Both models are trained online using the Adam optimizer with a learning rate of $0.001$ and a batch size of $32$. Training is performed at each round using the most recent 32 observed examples. \cref{fig:all-bandits}(bottom) is a result with a deeper model.

\paragraph{Baselines and tuning.}
We compare against several baselines including $\varepsilon$-greedy, NeuralUCB~\cite{neuralucb}, NeuralTS~\cite{neuralts}, and FTPL~\cite{kveton2020randomized}. To accelerate training for NeuralUCB and NeuralTS, we approximate the confidence matrix inverse using only the reciprocals of the diagonal entries. In our proposed algorithm, $\deepfp$, instead of perturbing the full $kd$-dimensional parameter space, we selectively perturb only the $d$-dimensional subspace corresponding to the chosen arm. This masking avoids interference from other arms, ensuring more accurate value estimation for the selected action.

\subsection{Additional experiments}
We provide additional experiments by varying the norm of the true parameter $S$ and the cardinality of the context set $|\Ccal|$. The specific configurations are provided in each figure caption.

To further validate our algorithm in a neural contextual bandit setting, we conduct additional experiments using the MNIST~\cite{MNIST} and Fashion-MNIST~\cite{FashionMNISTAN} datasets. Each instance is represented as a $28 \times 28$ grayscale image with label set size $K=10$, naturally forming a $K$-armed classification bandit problem. At each round, the agent receives a context composed of $K$ image tensors, where the $i$-th arm corresponds to the image being placed in the $i$-th channel slot, and all others set to zero. This yields a $K \times 28 \times 28$ tensor input for each arm. The model $f$ used to estimate the expected reward is a shared convolutional neural network that takes the entire $K$-channel tensor as input and outputs a scalar score for each arm, promoting parameter sharing across arms.

The architecture of the model is as follows: two convolutional layers are applied with ReLU activation and $2 \times 2$ max pooling after each. The first layer uses $32$ filters and the second $64$ filters, both with kernel size $3 \times 3$ and stride 1. The resulting $64 \times 7 \times 7$ feature map is flattened and passed through two fully connected layers with hidden size 128, followed by ReLU and a final output scalar. This design enables efficient learning of visual features while maintaining compatibility with the bandit framework through arm-wise shared representation.

\begin{figure*}[ht]
  \centering
  \includegraphics[width=\textwidth]{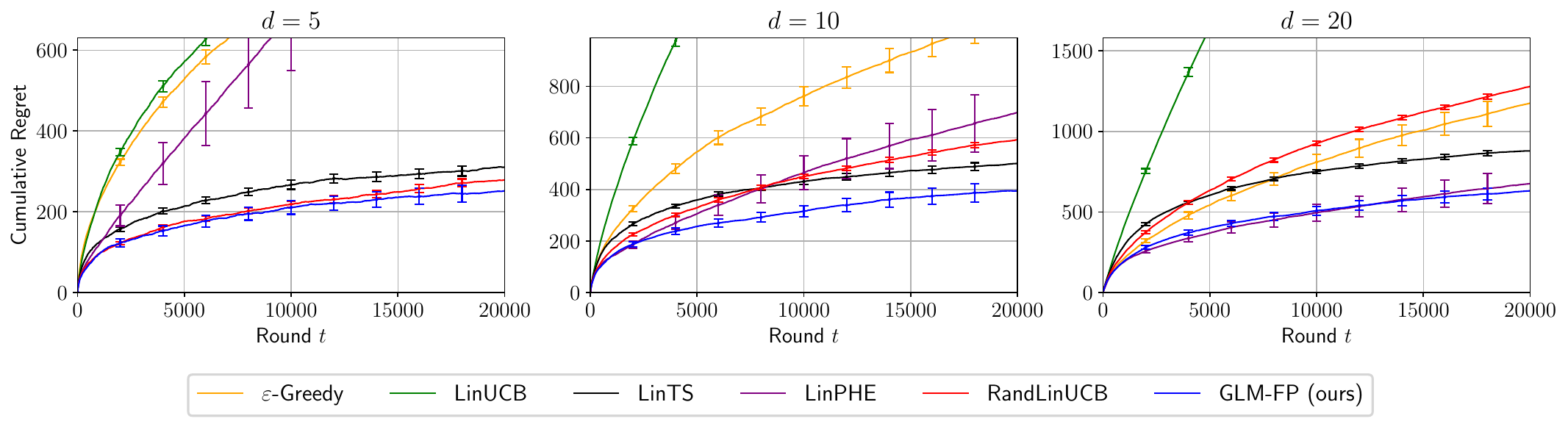}
  \caption{Linear Bandit. $|\Ccal|=1$, $d=\{5,10,20\}$, $K=100$, $S=1$.}
  \label{fig:linear_fixed_alltime}
\end{figure*}

\begin{figure*}[ht]
  \centering
  \includegraphics[width=\textwidth]{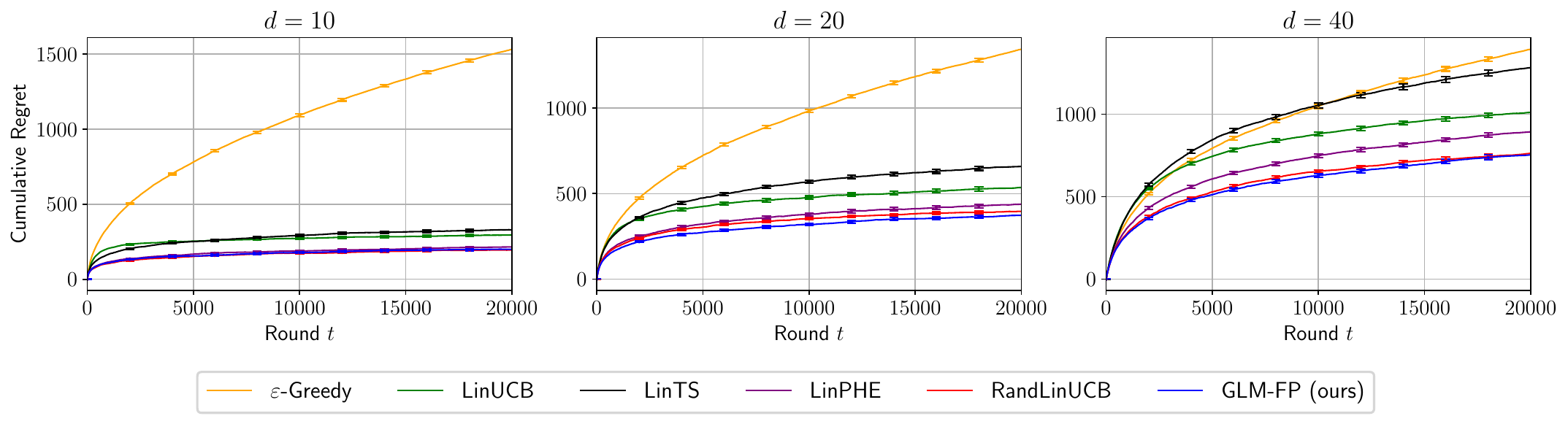}
  \caption{Linear Bandit. $|\Ccal|=T$, $d=\{10,20,40\}$, $K=100$, $S=2$.}
  \label{fig:linear_contextual_alltime_2}
\end{figure*}

\begin{figure*}[ht]
  \centering
  \includegraphics[width=\textwidth]{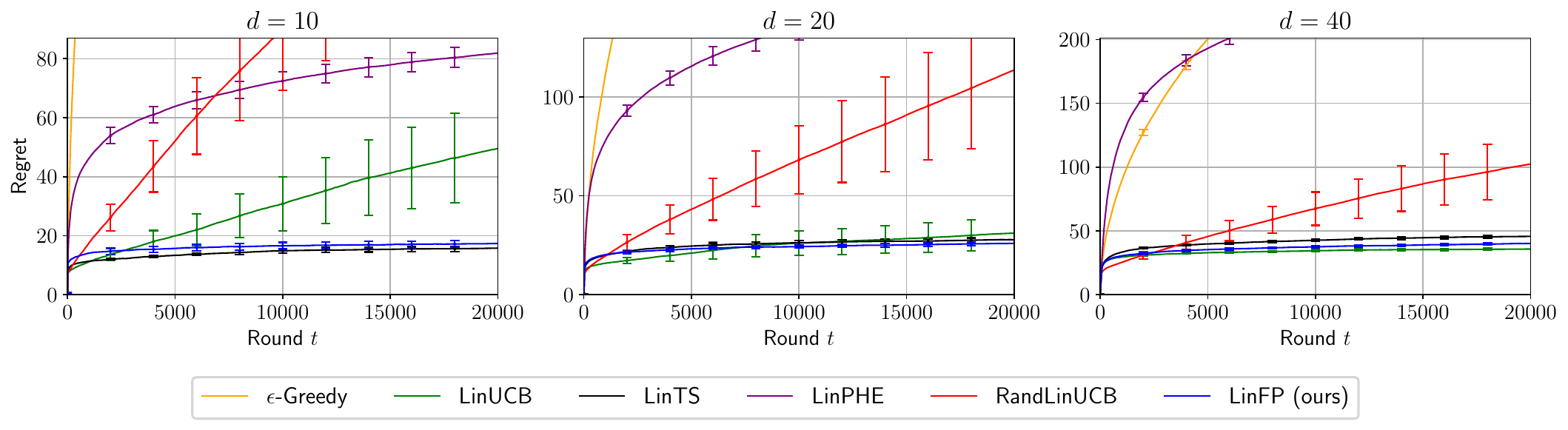}
  \caption{Linear Bandit with noise $\mathcal N(0,0.1^{2})$. $|\Ccal|=T$, $d=\{10,20,40\}$, $K=100$, $S=2$.}
  \label{fig:linear_noise01}
\end{figure*}

\begin{figure*}[ht]
  \centering
  \includegraphics[width=\textwidth]{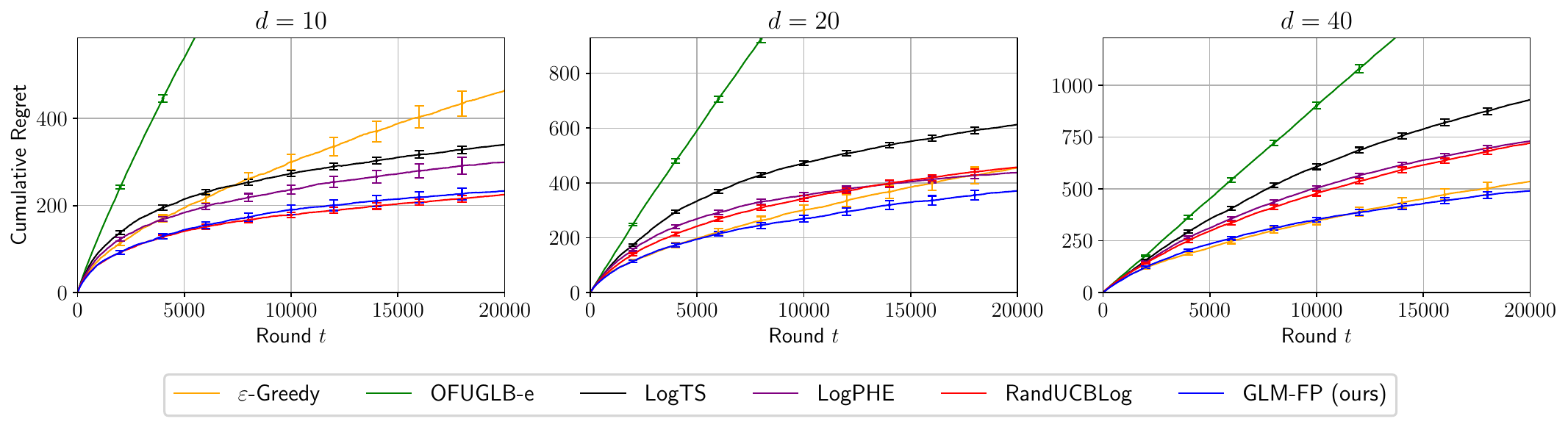}
  \caption{Logistic Bandit. $|\Ccal|=1$, $d=\{5,10,20\}$, $K=100$, $S=1$.}
  \label{fig:logistic_fixed_1}
\end{figure*}

\begin{figure*}[ht]
  \centering
  \includegraphics[width=\textwidth]{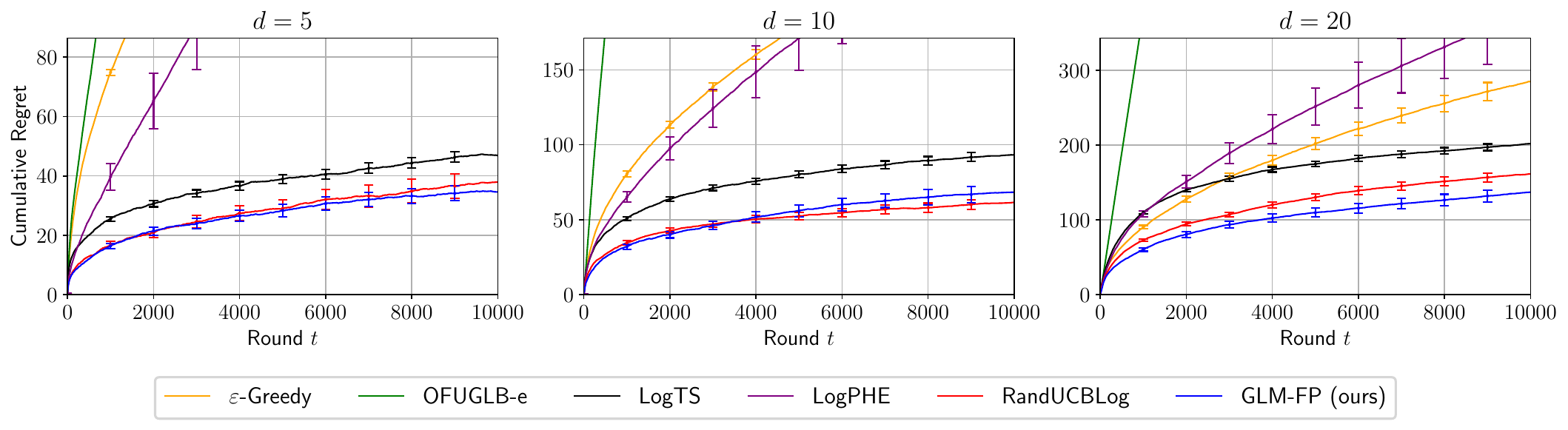}
  \caption{Logistic Bandit. $|\Ccal|=1$, $d=\{5,10,20\}$, $K=100$, $S=4$.}
  \label{fig:logistic_fixed_4}
\end{figure*}

\begin{figure*}[ht]
  \centering
  \includegraphics[width=\textwidth]{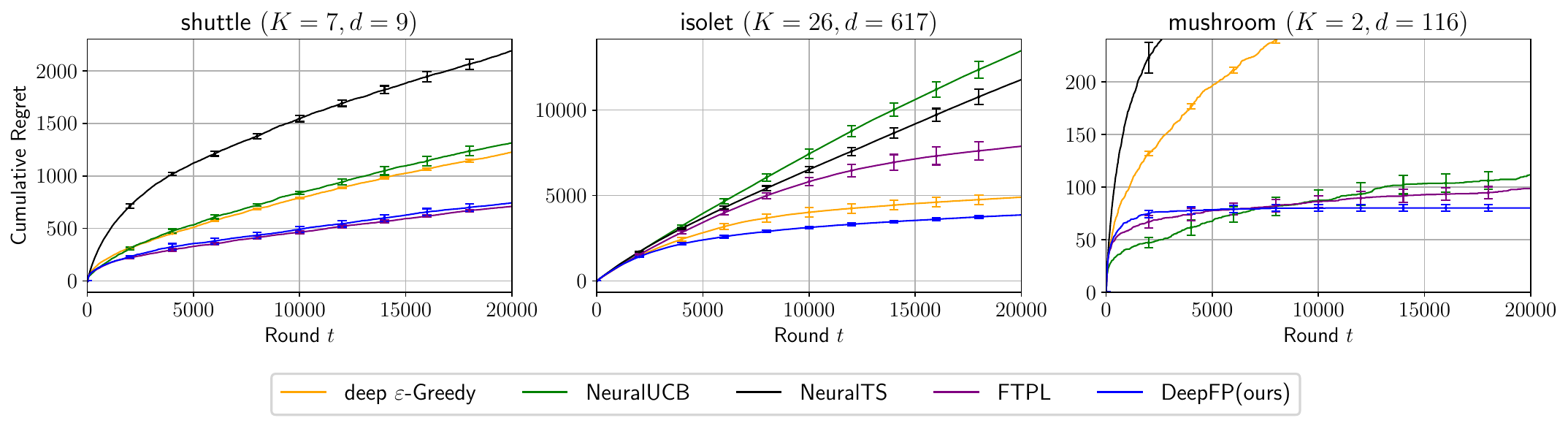}
  \caption{Neural Bandit. Multi-layer perceptron model with one hidden layer and output layer.}
  \label{fig:neural_single20000}
\end{figure*}

\begin{figure*}[ht]
  \centering
  \includegraphics[width=0.75\textwidth]{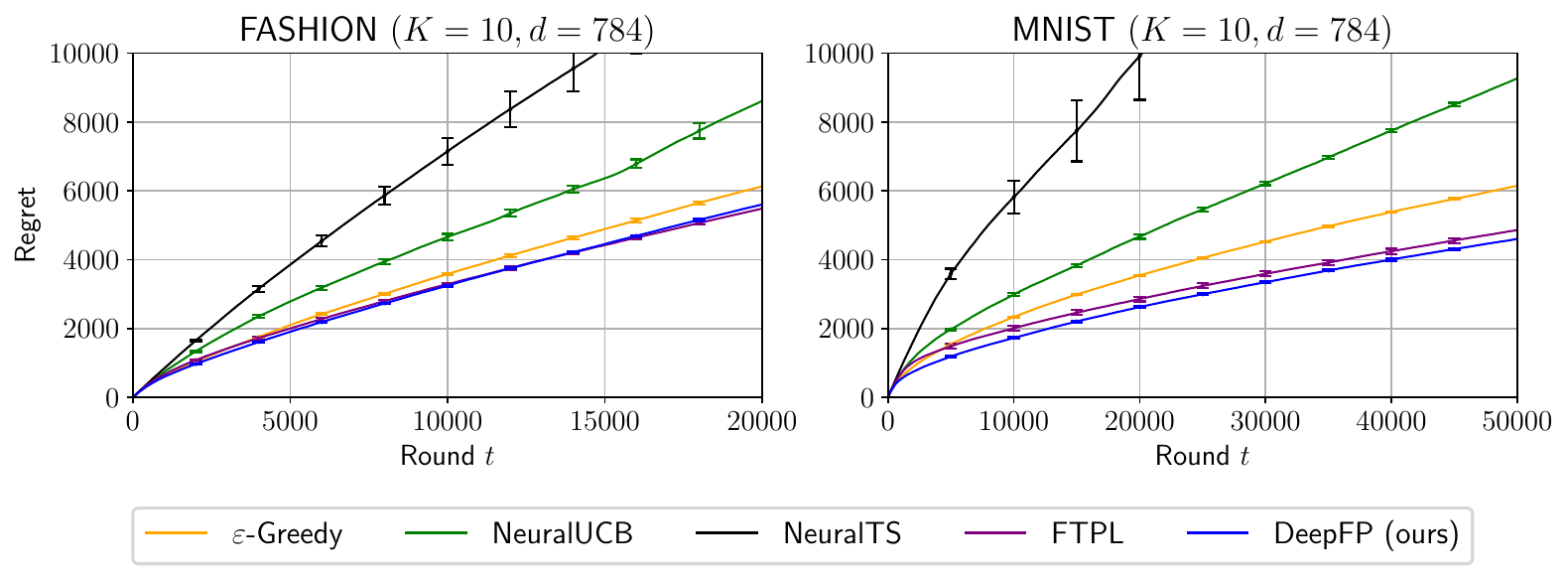}
  \caption{Neural Bandit. Experiments on MNIST dataset with a CNN model.}
  \label{fig:neural_mnist}
\end{figure*}

\end{document}